\documentclass[10pt,twocolumn,letterpaper]{article}

\usepackage{iccv}
\usepackage{times}
\usepackage{epsfig}
\usepackage{subfig}
\usepackage{graphicx}

\usepackage{amsmath}
\usepackage{amssymb}

\usepackage{float}
\usepackage{makecell}
\usepackage{caption}
\usepackage{booktabs}
\usepackage[table]{xcolor}

\usepackage{multirow}
\usepackage{tabularx,verbatim}
\usepackage{xspace}
\usepackage{setspace}
\usepackage{colortbl}
\usepackage{color, xcolor} 

\definecolor{mygray}{gray}{.9}
\definecolor{honeydew}{rgb}{0.94, 1.0, 0.94}
\definecolor{magicmint}{rgb}{0.67, 0.94, 0.82}
\definecolor{lightcyan}{rgb}{0.88, 1.0, 1.0}
\definecolor{lightgreen}{rgb}{0.86, 0.82, 1.0}

\usepackage[pagebackref=false, breaklinks=true, letterpaper=true, colorlinks, urlcolor=gray,
            citecolor=citecolor, linkcolor=linkcolor, bookmarks=false]{hyperref}
\definecolor{citecolor}{HTML}{0071BC}
\definecolor{linkcolor}{HTML}{ED1C24}

\colorlet{dark-green}{green!80!black}
\usepackage{pifont}
\newcommand{\cmark}{\textcolor{dark-green}{\ding{51}}}%
\newcommand{\xmark}{\ding{55}}%

\newcolumntype{*}{>{\global\let\currentrowstyle\relax}}
\newcolumntype{^}{>{\currentrowstyle}}
\newcommand{\rowstyle}[1]{\gdef\currentrowstyle{#1}#1\ignorespaces}

\definecolor{dt}{gray}{0.7}  %
\definecolor{lightgreen}{HTML}{D8ECD1}

\iccvfinalcopy 

\def\iccvPaperID{1747} 

\begin{document}


\title{Revisit Parameter-Efficient Transfer Learning: A Two-Stage Paradigm}
\author{Hengyuan Zhao$^1$,
Hao Luo$^2$,
Yuyang Zhao$^3$,
Pichao Wang$^2$,
Fan Wang$^2$,
Mike Zheng Shou$^1$\\
$^1$Show Lab, National University of Singapore, $^2$Alibaba Group, $^3$ National University of Singapore\\
{\tt\small (hengyuan.z, yuyang.zhao)@u.nus.edu, (michuan.lh, fan.w)@alibaba-inc.com} \\ 
{\tt\small (pichaowang, mike.zheng.shou)@gmail.com}
}

\maketitle

\begin{abstract}

    Parameter-Efficient Transfer Learning (PETL) aims at efficiently adapting large models pre-trained on massive data to downstream tasks with limited task-specific data.  
    In view of the practicality of PETL, previous works focus on tuning a small set of parameters for each downstream task in an end-to-end manner while rarely considering the task distribution shift issue between the pre-training task and the downstream task.
    In this paper, we propose a novel two-stage paradigm, where the pre-trained model is first aligned to the target distribution, and then the task-relevant information is leveraged for effective adaptation.
    Specifically, the first stage is to narrow the task distribution shift by tuning the scale and shift in the LayerNorm layers. In the second stage, to efficiently learn the task-relevant information, we propose a Taylor expansion-based importance score to identify task-relevant channels for the downstream task and then only tune such a small portion of channels, making the adaptation to be parameter-efficient.
    Overall, we present a promising new direction for PETL, and the proposed paradigm achieves state-of-the-art performance on the average accuracy of 19 downstream tasks. Codes will be available \href{https://github.com/showlab/TTC-Tuning}{here}. 

\end{abstract}

\section{Introduction}

%
Large vision transformer models~\cite{dosovitskiy2020image,liu2021swin,pvt} have demonstrated exceptional performance on large-scale image classification tasks~\cite{deng2009imagenet}.
Inspired by the successful usage of large language models~\cite{devlin2018bert,brown2020language,raffel2020exploring,lepikhin2020gshard}, there is a growing interest in leveraging the pre-trained knowledge from large vision transformer models for downstream tasks.
The most common and direct way is to fine-tune the whole model on the small downstream dataset.
Nevertheless, fine-tuning all the parameters (\textit{aka} full fine-tuning) on a small dataset can lead to two severe challenges: (1) full fine-tuning is prone to overfitting when the tuned massive weights of pre-trained models are not comparable with the limited downstream training data; (2) the high computation costs and storage requirements of a large number of model parameters (since each task requires storing a separate model) make it harder to be applied to extreme storage-constrained devices.

\begin{figure}[t]
\begin{center}
\includegraphics[width=0.86\linewidth]{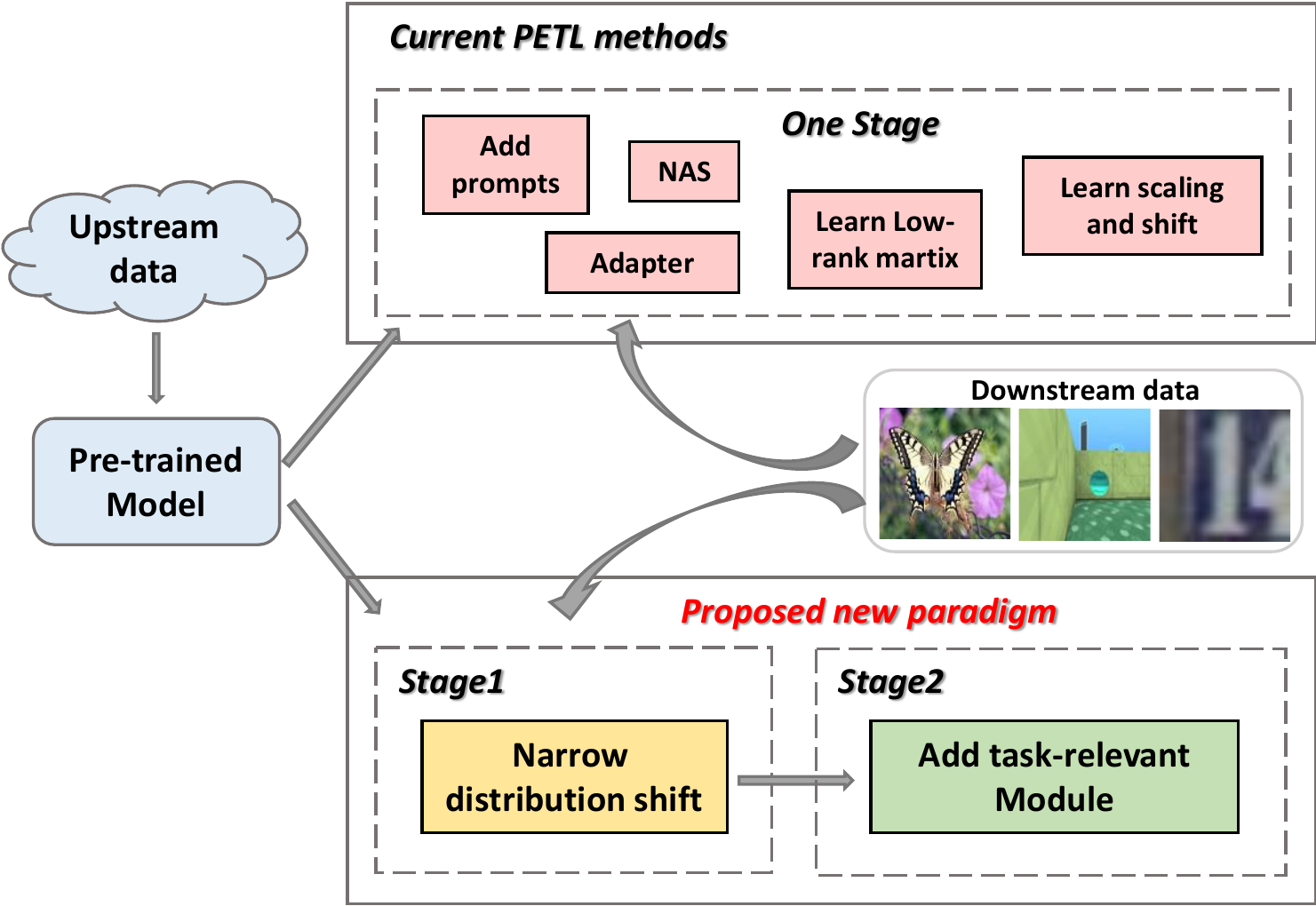}
\end{center}
 \vspace{-0.2in}
\caption{An illustration of our new paradigm.}
\label{fig:paradigm}
\vspace{-0.2in}
\end{figure}

To address the above two problems, recent research works investigate parameter-efficient transfer learning (PETL)~\cite{vpt,adapter,SSF}, aiming at efficiently adapting large models to downstream tasks with limited data. 
Mainstream PETL methods can be categorized into two types:
(1) designing additional modules (adapter~\cite{adapter} or visual prompt~\cite{vpt}) to learn task-relevant information; (2) narrowing the task distribution shift between the pre-trained and downstream tasks via feature scaling and shifting~\cite{SSF}.
Inspired by the effectiveness of the two approaches, we revisit PETL from the perspective of both task distribution shift and add the task-relevant module and present a novel two-stage paradigm called TTC-Tuning in this paper.

\begin{figure}[t]
\begin{center}
 \vspace{-0.2in}
\includegraphics[width=0.8\linewidth]{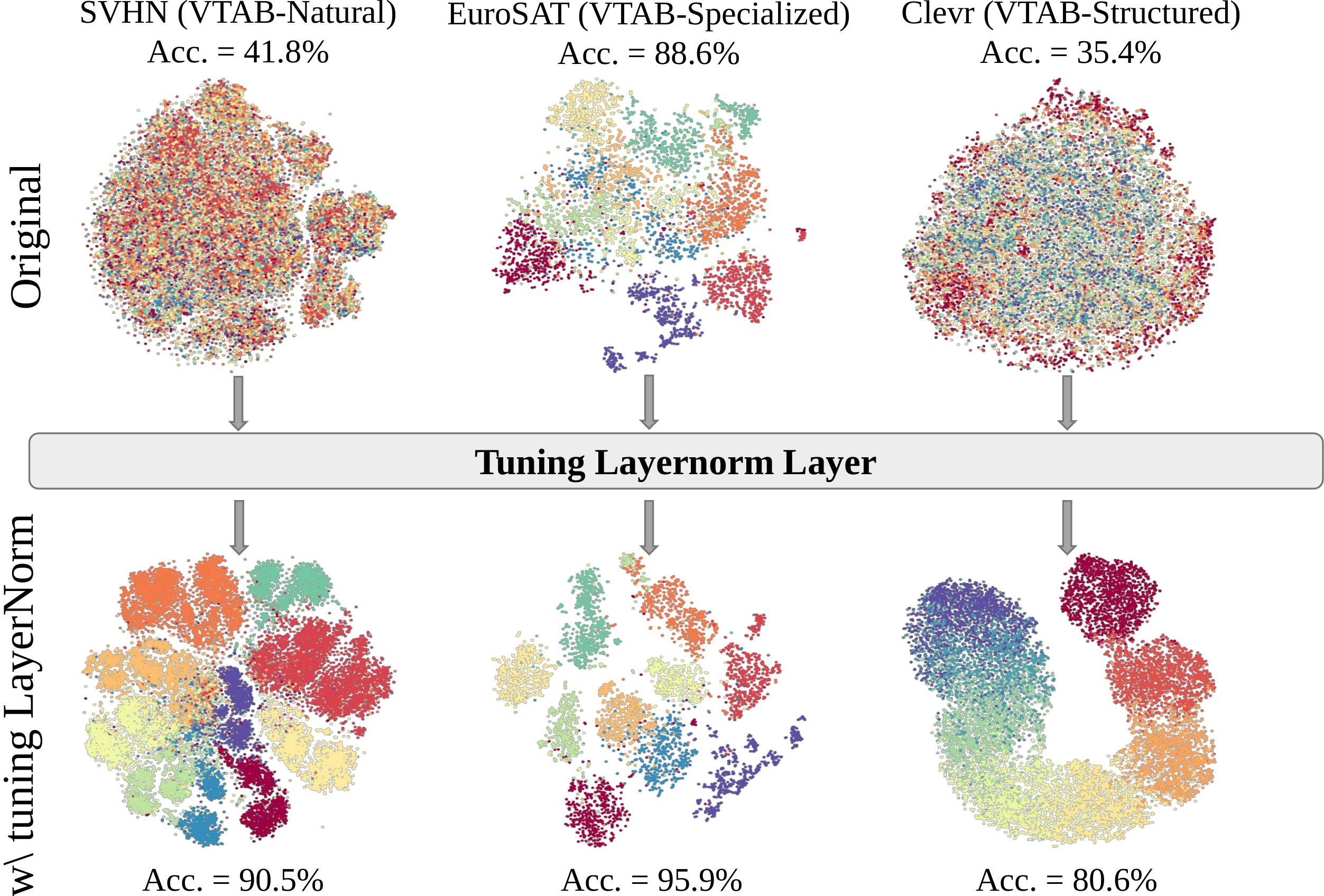}
\end{center}
 \vspace{-0.2in}
\caption{The t-SNE visualization of final \textsc{[CLS]} token of the test set from SVHN, EuroSAT, and Clevr tasks. ``Original" represents the feature extracted from the original backbone while ``w/ LayerNorm" tuning means extracted from the backbone tuned with LayerNorm only.}
\label{fig:distribution}
\vspace{-0.2in}
\end{figure}

For narrowing task distribution shift, SSF~\cite{SSF} inserts additional scale and shift parameters into MLP, MHSA and LayerNorm components to modulate the features. 
Instead of using additional parameters, tuning normalization layers is a common way to align distributions in transfer learning tasks~\cite{wang2020tent}. Thus, we follow the concept of modulating features but propose a more effective and efficient technique to align the task distribution, \ie, tuning the layer normalization (LayerNorm) parameters.
As shown in Fig.~\ref{fig:distribution}, LayerNorm tuning can greatly improve the discrimination ability of the pre-trained model on downstream tasks.
Compared with SSF~\cite{SSF}, LayerNorm tuning uses less than 15\% parameters (0.03M \textit{v.s.} 0.21M) but outperforms SSF by 0.5\% in the absolute top-1 accuracy.
Therefore, we adopt LayerNorm tuning as our first step for task distribution alignment.


\begin{figure*}[t]
\begin{center}
\vspace{-0.4in}
\includegraphics[width=0.8\linewidth]{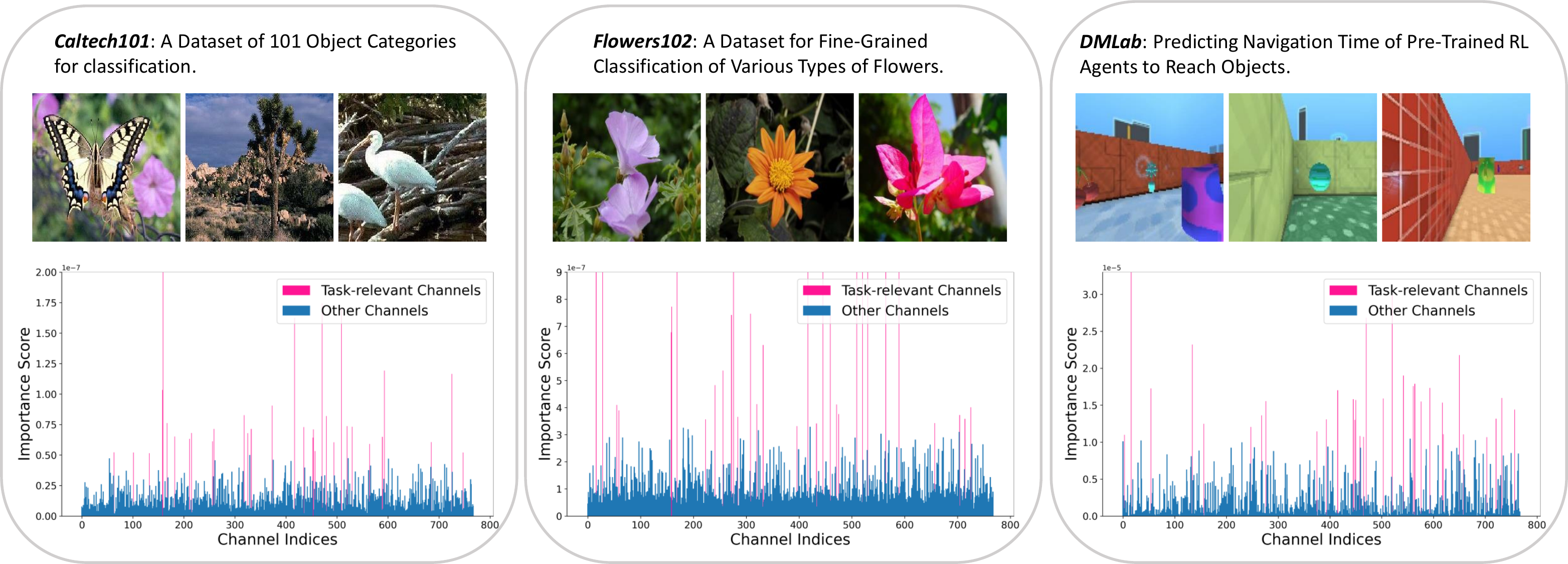}
\end{center}
 \vspace{-0.1in}
\caption{We have identified task-relevant channels from the last layer of ViT-B on three tasks. This suggests that different tasks may prioritize different channels within the same layer.}
\label{fig:ic-show}
 \vspace{-0.2in}
\end{figure*}

Besides aligning the task distribution, TTC-Tuning also considers adding task-relevant module which has been shown to be crucial by previous studies. 
For some challenge datasets, such as medical images (Camelyon dataset~\cite{Veeling2018qh}) and 3D scene images (Clevr-Count~\cite{johnson2017clevr}), only aligning the task distribution lead to inferior improvement due to the large knowledge gap between the downstream task and pre-trained model. Previous PETL methods~\cite{vpt, adapter,lora, jie2022convolutional, noah, chen2022adaptformer, SSF} mainly propose parameter-efficient tuning modules to implicitly leverage the task-relevant information by adding tokens or adapting the whole features. However, these methods treat each parameter as equivalent and just insert some fixed modules to automatically adapt the whole network to the downstream tasks. Here, we raise an essential question: 
\textit{can we identify the important parameters for a specific downstream task and then fine-tune only these task-relevant parameters?}

Inspired by the channel bias in few-shot learning~\cite{luo2022channel} and model pruning~\cite{li2016pruning}, we hypothesize and experimentally validate that channel inequality exists in different tasks. 
We can explicitly leverage such task-relevant information to tune only a small portion of task-relevant channels, leading to comparable or even better performance.
To verify this hypothesis, we investigate the contributions of individual channels in downstream adaptation. The contributions are measured by a proposed Taylor expansion-based importance score.
As shown in Fig.~\ref{fig:ic-show}, different tasks have different task-relevant channels in the same layer. 
Thus, we can select the task-relevant channels based on the contributions and utilize a simple adapter to transform such channels for efficient adaptation. 

In summary, our contributions are three-fold:
\begin{itemize}
    \item We propose a new two-stage paradigm to solve the PETL from the perspective of both task distribution shift and add task-relevant tunable module.
    \item We experimentally verify the effectiveness of only tuning the LayerNorm layer to align distributions and develop a novel tuning module that first selects task-relevant channels via the proposed Taylor expansion-based importance score. Such designs lead to a few extra parameters.
    \item Our novel paradigm outperforms the previous state-of-the-art method SSF~\cite{SSF} with a 1.7\% increase in accuracy across 19 downstream tasks. This result highlights the effectiveness of our approach and its potential to make a significant impact in various applications.
\end{itemize}



\section{Related Work}
\raggedbottom

\subsection{Vision Transformers}

Transformers \cite{vaswani2017attention} have shown remarkable performance on natural language processing and computer vision tasks. Numerous vision transformers \cite{chen2021crossvit,d2021convit,dong2022cswin,ali2021xcit,fan2021multiscale,han2021transformer,rao2021dynamicvit,yuan2021tokens,touvron2021going,liu2021swin,wang2021kvt,zhou2021elsa} have been proposed following the pioneering work of ViT \cite{dosovitskiy2020image}. Most of these models gradually increase in size to achieve state-of-the-art results and learn rich representations through various architectural designs. 
Adopting these models for downstream tasks significantly reduces the training complexity and delivers promising results rapidly. Given a plain Vision Transformer (ViT)~\cite{dosovitskiy2020image} with $L$ layers and an input image $I \in \mathbb{R}^{3 \times H \times W}$ that first divided into $N$ non-overlapped patches and then passed into an embedding layer projected into $D$ dimensions. Each transformer layer includes a multi-head self-attention block (MHSA) and a multi-perceptron block (MLP). 

\subsection{Parameter-Efficient Transfer Learning}

PETL focuses on adapting the pre-trained model on a downstream task with a few parameters. 
Two lines of PETL approaches have been proposed recently. On the one hand, learning task-relevant information by applying prompts \cite{vpt,liu2022prompt,xing2022class,zheng2022prompt,nie2022pro,wang2022fine} to the input tokens or adding a trainable module \cite{adapter, chen2022adaptformer, jie2022convolutional,chen2022conv, zhang2023multimodal} to adapt pre-trained information have acquired promising results for the performance and efficiency. On the other hand, aligning the distribution between pre-trained and downstream tasks has been shown to be a strong baseline, as demonstrated in~\cite{SSF}.

\noindent \textbf{Task-relevant modules.}
\textbf{VPT}~\cite{vpt} injects the prompts into the transformer layer's input tokens with a small number of extra parameters. However, one main limitation of VPT is that it relies on hand-crafted selection to determine the optimal prompt length for each task. This can be inflexible when applying the method to new tasks. VPT includes two variants VPT-Shallow and VPT-Deep associated with the number of inserted layers. VPT-Shallow only inserts prompts into the first transformer layer $L_1$ and VPT-Deep inserts all the transformer layers. Given the input tokens $x \in \mathbb{R}^{(N + 1) \times D}$ and the prompts $P \in \mathbb{R}^{n \times D}$ that contains $n$ prompts with dimension $D$, we can formulate the combined tokens $x^{\prime}$ is 
\begin{equation}
    x^{\prime} = [x;P],
    \label{eqn:eq2}
\end{equation}
where $x^{\prime} \in \mathbb{R}^{(N+n+1) \times D}$ will be passed into the following MHSA and MLP blocks.

\textbf{Adapter}~\cite{adapter} proposes an MLP-like module, a successful design that 
adopts a residual pathway to keep the original information and transform task-relevant information by learning a down-projection $W_{down} \in \mathbb{R}^{D^{\prime} \times D}$ (where $D^{\prime} \ll D$) and an up-projection $W_{up} \in \mathbb{R}^{D \times D^{\prime}}$ with a nonlinearity activation operation $\Phi$. Given an input tokens $x^l \in \mathbb{R}^{(N + 1) \times D}$ in $l$-th layer, the output of a adapter block is
\begin{equation}
    x_{out}^l = x^l + [W^l_{up}\Phi(W^l_{down}[x^l]^T)]^T,
    \label{eqn:eq1}
\end{equation}
where $[ \cdot ]^T$ represents transpose operation. However, the number of trainable parameters in Adapter-like methods is not small and produces inferior performance.
Besides, LoRA \cite{lora} optimizes a low-rank decomposition matrix with a low intrinsic dimension to project the matrices of query, key, and value used in the MHSA block in ViT. Furthermore, a neural architecture search algorithm called NOAH \cite{noah} has been proposed, which incorporates Adapter~\cite{adapter}, LoRA~\cite{lora}, and VPT~\cite{vpt} into its network search space.

\noindent \textbf{Narrow task distribution shift.} \textbf{SSF}~\cite{SSF} In addition to the above prompt-based and adapter-based methods, a recently introduced technique called SSF that has shown promising results involves scaling and shifting the features of the pre-trained model. SSF~\cite{SSF} leverages two learnable vectors $\gamma \in \mathbb{R}^D$ and $\beta \in \mathbb{R}^D$ to scale and shift the feature map in each transformer operation (\ie, Linear operation or LayerNorm operation). Assuming the input of SSF module is $x \in \mathbb{R}^{(N+1) \times D}$, the output $y \in \mathbb{R}^{(N+1) \times D}$ can be written as following
\begin{equation}
    y = \gamma \ast x + \beta,
    \label{eqn:eq3}
\end{equation}
where $\ast$ is the Hadamard product. Motivated by this work, we extend this method to tuning the LayerNorm layer to reduce the distribution shifts and demonstrate the effectiveness on multi-downstream tasks.

\section{Approach}

\begin{figure*}[t]
\begin{center}
\vspace{-0.2in}
\includegraphics[width=0.82\linewidth]{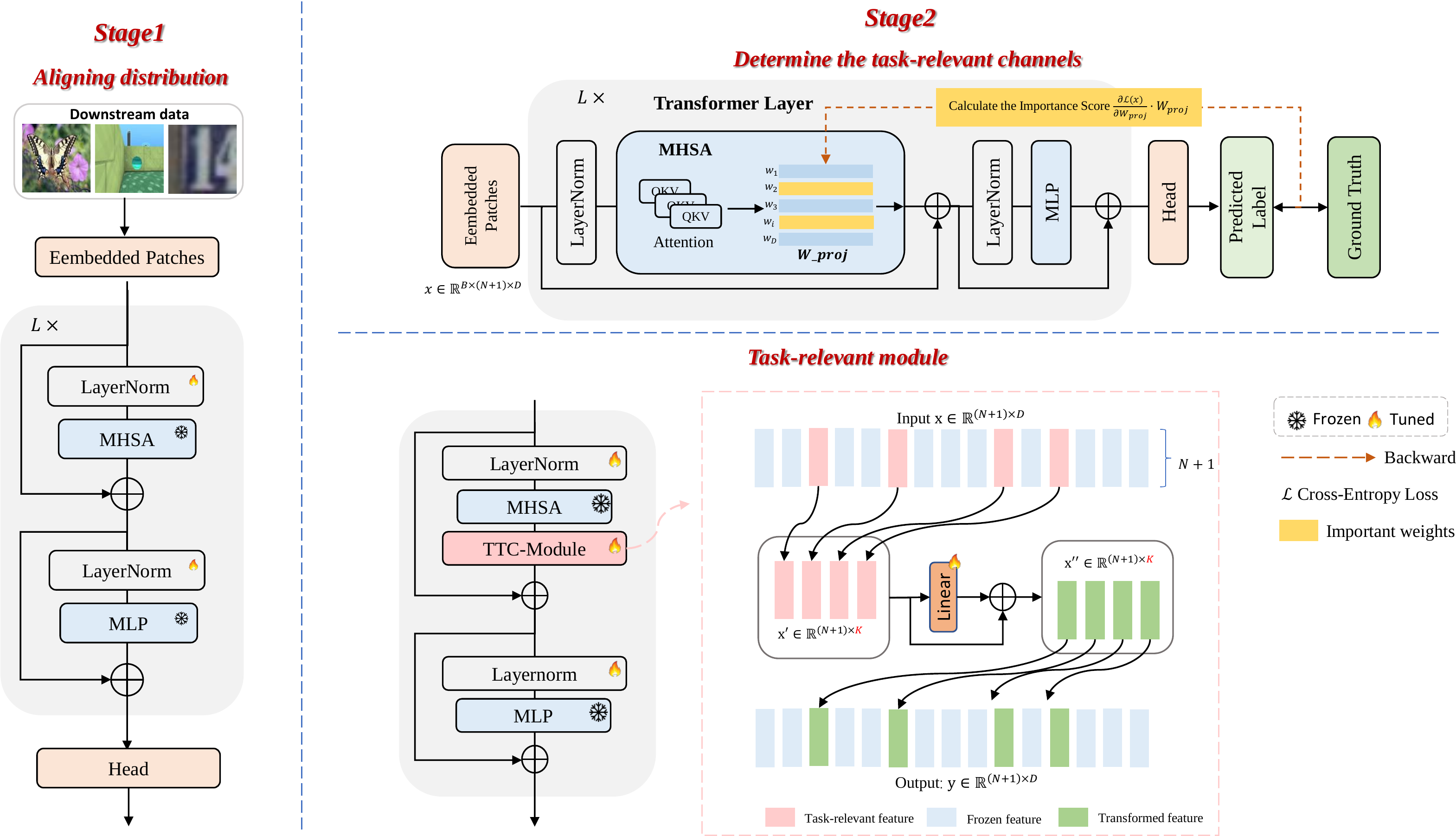}
\end{center}
 \vspace{-0.2in}
\caption{An overview of our novel paradigm to parameter-efficient transfer learning.}
\label{fig:overview}
\vspace{-0.2in}
\end{figure*}

We propose a two-stage paradigm for achieving parameter-efficient transfer learning, as shown in Fig.~\ref{fig:overview}. In the first stage, we align the task distribution by tuning the LayerNorm layer while keeping the other components of the original backbone frozen. In the second stage, we use a Taylor expansion-based Importance Score (TIS) to identify the most relevant channels for the downstream task, by computing gradients on the training set with stage1's model. Then, we introduce the TTC-Module, a tunable module that transforms the task-relevant channels while freezes other channels.

\begin{figure}[t]
\begin{center}
\includegraphics[width=0.48\linewidth]{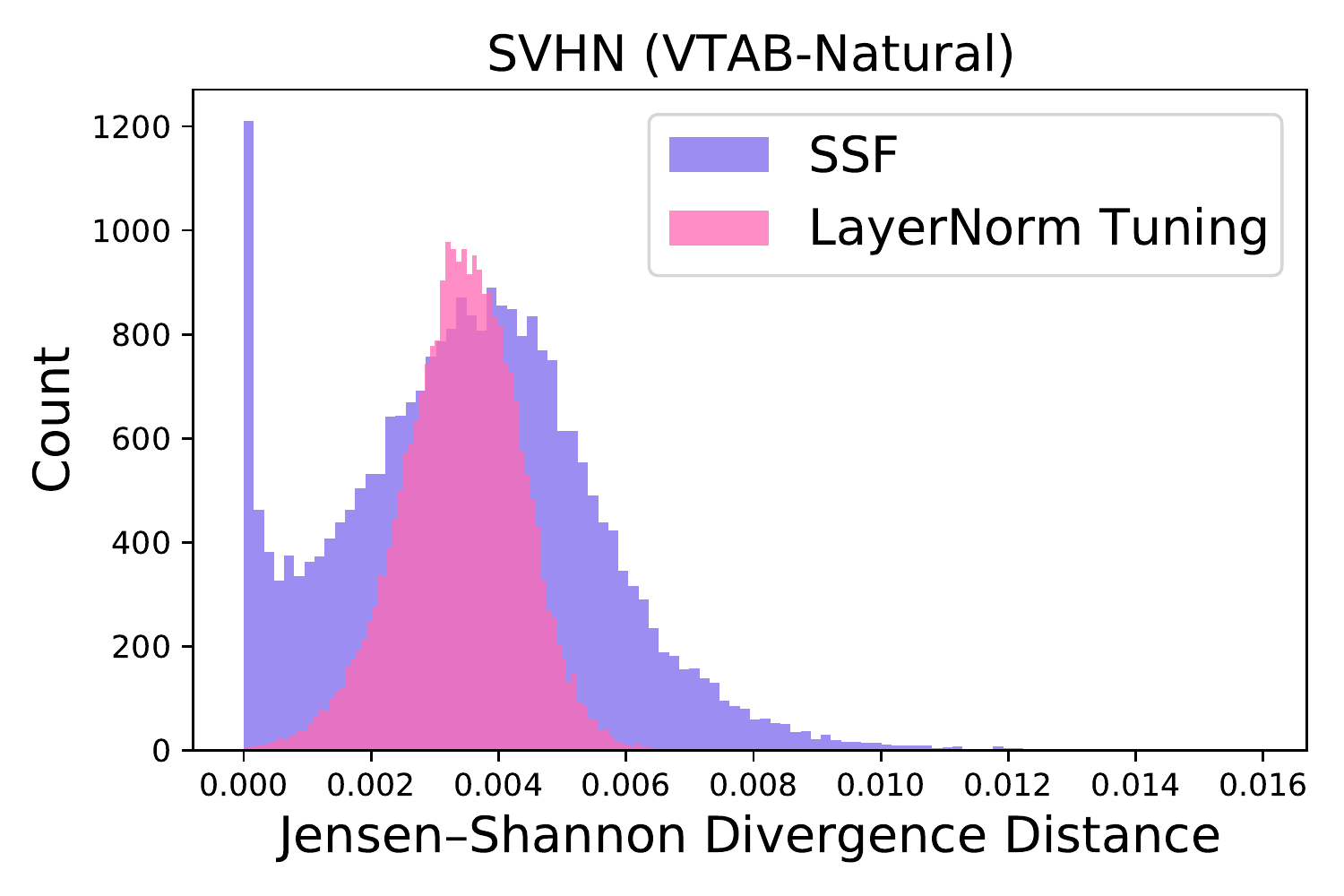}
\includegraphics[width=0.48\linewidth]{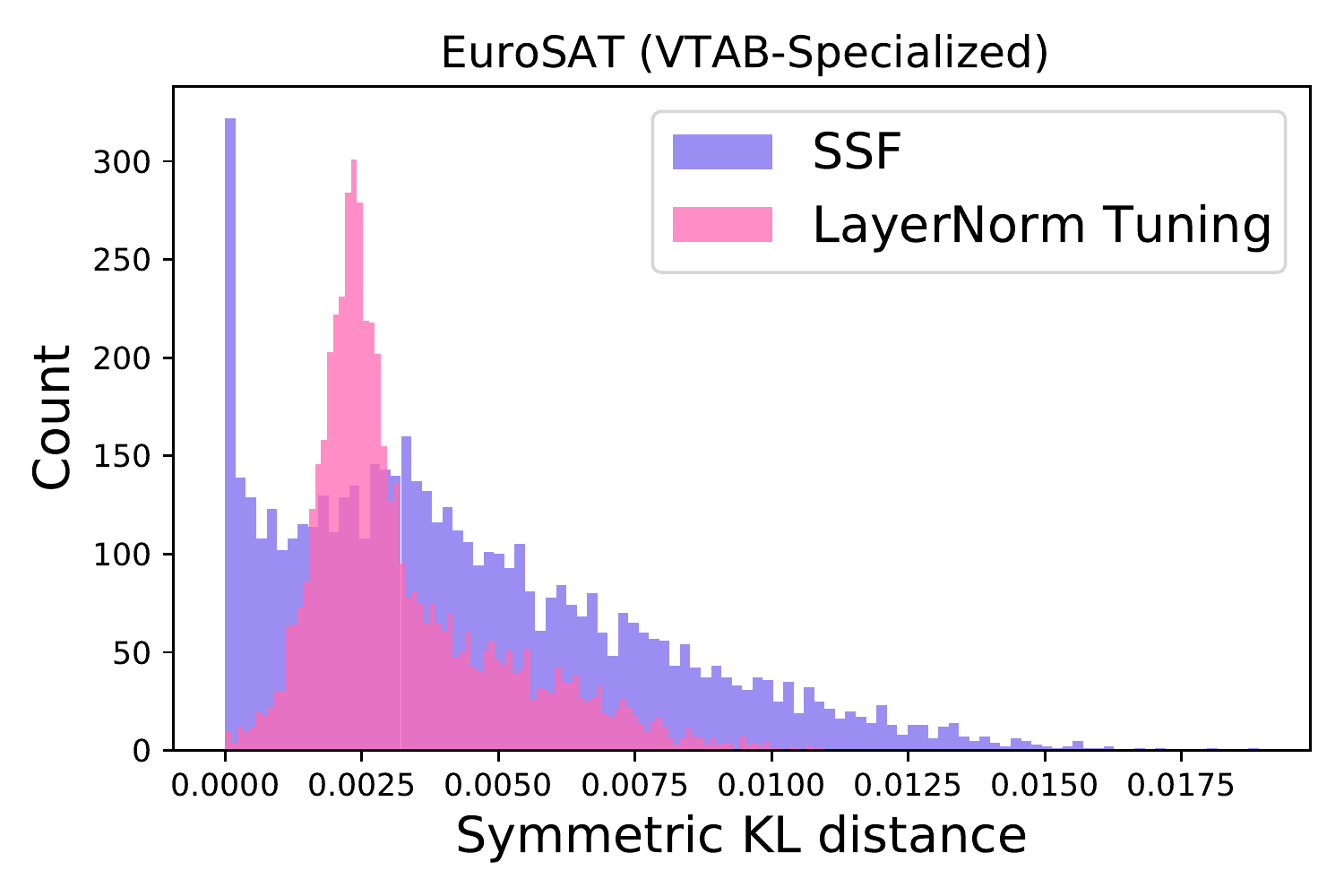}
\end{center}
 \vspace{-0.2in}
\caption{We analyzed the distribution of JSD for the \textsc{[CLS]} token on the SVHN and EuroSAT tasks. The JSD between the original feature and the feature generated by SSF is represented by ``SSF," while the JSD between the original feature and the feature produced after tuning the LayerNorm layer is represented by ``LayerNorm Tuning".}
\label{fig:KL-compare}
\vspace{-0.2in}
\end{figure}

\subsection{Narrow the Task Distribution Shift}
In this section, we first briefly review the Layer Normalization (LN)~\cite{ba2016layer}. LN is a widely used normalization technique in transformers~\cite{vaswani2017attention, dosovitskiy2020image} to solve the problem of the inconsistent amount of input tokens in natural language processing tasks and provide valid normalization in the MLP block. We empirically find that for PETL, tuning LayerNorm layer could efficiently change the mean and variance of feature distribution as mentioned in Fig.~\ref{fig:distribution}.
Assuming the input $x \in \mathbb{R}^{B \times (N +1) \times D}$, the output $y \in \mathbb{R}^{B \times (N +1) \times D}$ can be formulated as:
\begin{equation}
    y = \frac{x - E[x]}{\sqrt{\text{Var}[x]+\epsilon}} \ast \gamma + \beta,
    \label{eqn:eq4}
\end{equation}
where $\gamma$ and $\beta$ are scaling and bias factors, respectively. $E[\cdot]$ and $\text{Var}[\cdot]$ are expectations and variances that will lead to zero mean and unit variance.

Second, we analyze the statistics of the last \textsc{[CLS]} token to compare the effectiveness of tuning LN and SSF module. In particular, we assume the baseline as the original model distribution and compute the distances between this distribution with the distribution of LN tuning and SSF, respectively. Consider two probability distributions are $p$ and $q$, we use Jensen–Shannon Divergence (JSD)~\cite{endres2003new} as the metric to compute the distance $\mathcal{L}$ as:
\begin{equation}
\begin{split}
    \mathcal{L} &= \frac{1}{2}(\mathcal{KL}(\log(p), m) + \mathcal{KL}(\log(q), m)),\\
    m &= (p + q) / 2,
    \label{eqn:eq5}
\end{split}
\end{equation}
where $\mathcal{KL}$ is Kullback–Leibler divergence~\cite{kullback1951information}. 
Figure~\ref{fig:KL-compare} displays the distance distributions, where the blue histogram represents the distance between the original model and the model trained with SSF, while the pink histogram compares the distance of the original model with the LayerNorm model. Examining the JSD distribution, the range and the covariance of SSF are larger than LayerNorm. Notably, a significant number of samples are located at zero, indicating that the distribution is the same as the original model. On the other hand, a considerable number of samples are located far away from the original model, suggesting that SSF may fit some samples while ignoring others. In contrast, our LayerNorm tuning has a more compact distribution and appears unbiased towards any particular sample.

\subsection{Task-Relevant Channel Selection using Taylor Expansion-Based Importance Scores}


While aligning the distribution between pre-trained and downstream tasks can be effective for small distribution shifts, to handle various task distribution shifts, we need to introduce an extra learnable module that has been proved crucial by other PETL methods proposed~\cite{vpt,adapter,lora,noah}. However, unlike these methods treat each channel equally during fine-tuning, we hypothesize that only tuning a small portion of full channels is enough for adaptation. We note that the network weights are closely related to the task labels as mentioned in ~\cite{li2016revisiting}, and thus we aim to select task-relevant weights by feeding the downstream training set. Various methods~\cite{luo2017thinet,li2016pruning, he2018soft, wangvtc} for selecting network weights have been studied in the fields of network pruning and compression. To this end, we propose a Taylor expansion-based Importance Score (TIS) to evaluate the importance of each weight.

We conjecture that the task-relevant weights highly influence the network output, and removing these weights will drastically influence the loss value. Thus, the importance of weight can be quantified by the difference in loss induced by removing this weight. Given a subset $\{x,y\}$ randomly sampled from training set, the importance score $I_{w_i^j}$ of a weight parameter $w_i^j \in \mathbb{R}^{1 \times 1}$ can be formulated by
\begin{equation}
    I_{w_i^j} = (\mathcal{L}(\mathcal{F}(x, \mathcal{W}), y | w_i^j=0) - \mathcal{L}(\mathcal{F}(x, \mathcal{W}), y))^2,
    \label{eqn:eq6}
\end{equation}
where $\mathcal{L}$ is the task-specific loss (the cross-entropy loss in this paper), $\mathcal{F}$ is the transformer network, $\mathcal{W}$ is the total model weights and $y$ are the labels of data $x$. As previous studies~\cite{molchanov2019importance, wangvtc, yang2021nvit} point out that this score can be approximated with the first-order Taylor expansion. Thus, the final importance score $\hat{I}_{w_i^j}$ of a weight parameter $w_i^j$ can be rewritten as:
\begin{equation}
    \hat{I}_{w_i^j} = \frac{\partial{\mathcal{L}(x)}}{\partial{w_i^j}} \cdot w_i^j.
    \label{eqn:eq7}
\end{equation}
Thus the importance score $\hat{I}_{w_i^j}$ can be represented with a gradient term and the weight parameter $w_i^j$.

\begin{figure}[t]
\begin{center}
\includegraphics[width=0.65\linewidth]{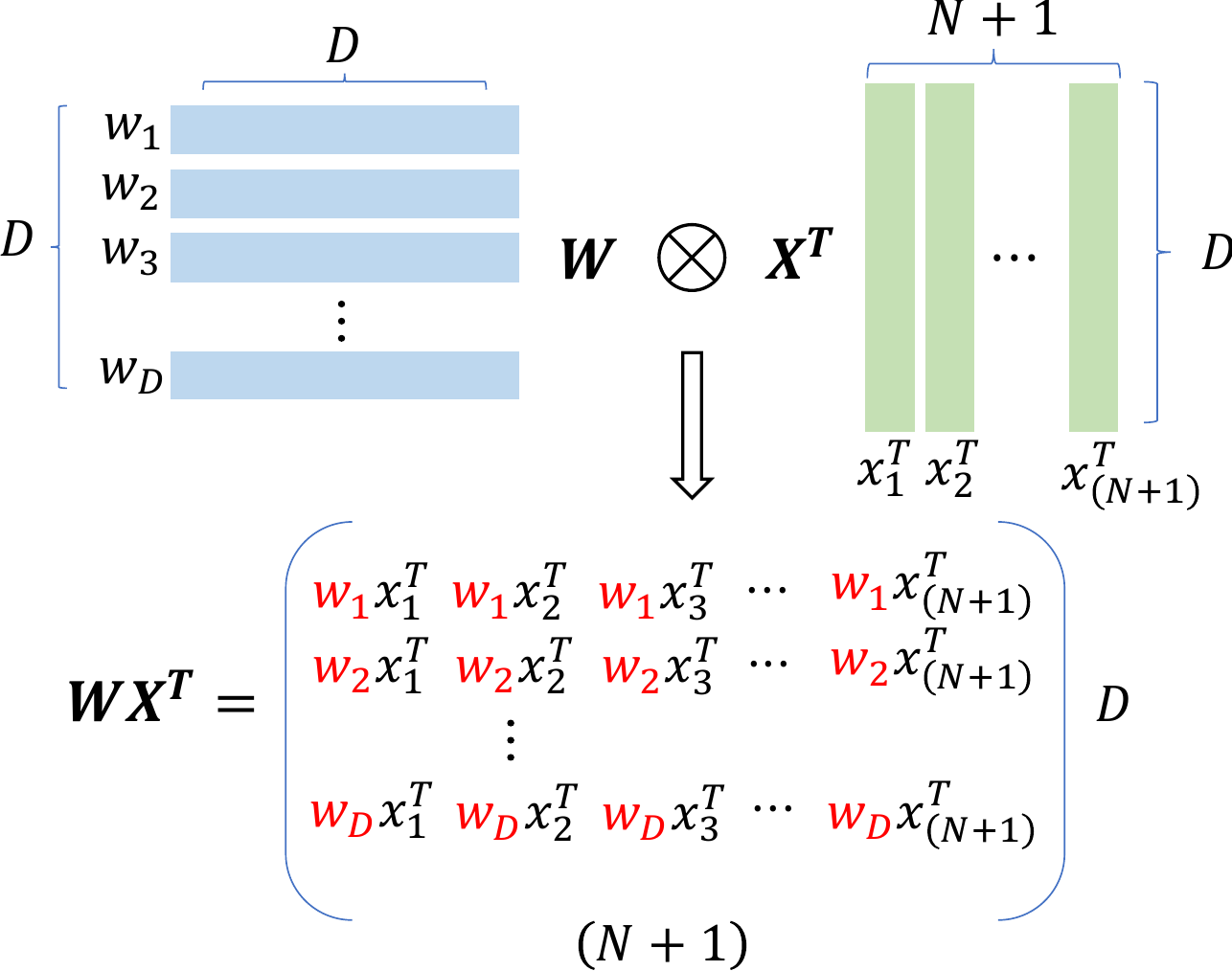}
\end{center}
 \vspace{-0.2in}
\caption{Illustration of the Decomposition of a Linear Operation.}
\label{fig:weightm}
\vspace{-0.2in}
\end{figure}

 Up to this point, we can use the above score to evaluate the task-relevant weights. However, our method aims to find task-relevant channels of a given feature map. Thus, we need to translate the task-relevant weights to task-relevant channels. As shown in Fig.~\ref{fig:weightm}, we first decompose the process of linear operation. Assuming a weight matrix $\textbf{W}\in \mathbb{R}^{D \times D}$ and a feature map $\textbf{X} \in \mathbb{R}^{(N+1) \times D}$, we can get the output $\textbf{Y}\in \mathbb{R}^{(N+1) \ \times D}$ as:
\begin{equation}
    \textbf{Y} = [\textbf{W}\textbf{X}^T]^T.
    \label{eqn:eq8}
\end{equation}
We define each weight $w_i \in \mathbb{R}^{1 \times D}$ in $\textbf{W}$ and each token $x^T_i \in \mathbb{R}^{D \times 1}$ in $\textbf{X}^T \in \mathbb{R}^{D \times (N+1)}$. Summing the items of output $\textbf{Y}$ in channel-wise we can get:
\begin{equation}
\begin{split}
    \text{Sum}(\textbf{Y}, dim=1) = [&w_1(x^T_1+x^T_2+...+x^T_{N+1}); \\
    &w_2(x^T_1+x^T_2+...+x^T_{N+1});\\
    & ...\\
    &w_D(x^T_1+x^T_2+...+x^T_{N+1});].
    \label{eqn:eq9}
\end{split}
\end{equation}
Thus, we can find that task-relevant weights $w_i$ could represent the task-relevant channels of a given feature map $\textbf{Y}$. Calculating the importance score of a weight $w_i$ could be approximated by summing over Eq.~\ref{eqn:eq7} of all the parameters in $w_i$, \ie, the final importance score $\mathcal{S}_i$ can be calculate as:
\begin{equation}
    \mathcal{S}_i = \sum_{j\in \mathcal{J}}\hat{I}_{w_i^j},
    \label{eqn:eq10}
\end{equation}
where $\mathcal{J}$ represents the index set of a weight $w_i$ and $w_i^j \in \mathbb{R}^{1 \times 1}$ is a parameter in $w_i$. 


\subsection{Task-Relevant Module}
\label{sec:adapt}

Having obtained Taylor expansion-based Importance score, we will select top-$K$ task-relevant channels of each feature map in transformer layers. Assuming a feature map is $x \in \mathbb{R}^{(N+1) \times D}$, we will select $K$ largest value of importance score vector $\mathcal{S}=[\mathcal{S}_1, \mathcal{S}_2, ..., \mathcal{S}_i \in \mathbb{R}^{1 \times 1}, 1\leq i \leq D]$ in this feature map. The selected feature $x^{\prime} \in \mathbb{R}^{(N+1) \times K}$ then be fed into a trainable linear layer and outputs the transformed feature
\begin{equation}
    x^{\prime\prime} = x^{\prime} + \mathrm{Linear}(x^{\prime}),
    \label{eqn:eq11}
\end{equation}
where $x^{\prime\prime} \in \mathbb{R}^{(N+1) \times K}$ will be passed into the next operation in the transformer layer and $\mathrm{Linear}(\cdot)$ is linear layer operation that the only new involved layer with parameters $K \times K$. Here, we adopt a shortcut connection to preserve the original information and prevent error accumulation across the transformer layers. This strategy helps alleviate the training difficulty.


\noindent\textbf{Tuning channels \vs Tuning weights.}
Note that if we consider $K$ task-relevant weights, the extra parameters will be $K \times D$, and the FLOPs will be $N \times K \times D$. On the other hand, focusing on $K$ task-relevant channels only requires a $K \times K$ linear layer for tuning, with the FLOPs count of $N \times K \times K$. In this paper, we set our default values as $K=96$ and $D=768$, and the number of extra parameters for tuning task-relevant weights is 8 times larger than that for tuning task-relevant channels. Therefore, to maintain fewer extra parameters for storage, it is better to tune task-relevant channels instead. We also attempted to tune the task-relevant weights directly, but the results for 19 downstream tasks were inferior compared to tuning the task-relevant channels as illustrated in Tab.~\ref{tab:vit-vtab-weights-channels}. We hypothesize that the linear combination of task-relevant channels will contribute more to task performance.




\begin{table*}[t]
\vspace{-0.1in}
\begin{center}
\large
\resizebox{0.92\linewidth}{!}{
\begin{tabular}{lc|ccccccc|cccc|cccccccc|c}
\multirow{2}{*}{} &  & \multicolumn{7}{c}{\textbf{Natural}} & \multicolumn{4}{c}{\textbf{Specialized}} & \multicolumn{8}{c}{\textbf{Structured}} \\
& \rotatebox{90}{\# Params (M)} & \rotatebox{90}{Cifar100} & \rotatebox{90}{Caltech101}  & \rotatebox{90}{DTD} & \rotatebox{90}{Flower102} & \rotatebox{90}{Pets} & \rotatebox{90}{SVHN} & \rotatebox{90}{Sun397} & \rotatebox{90}{Camelyon} & \rotatebox{90}{EuroSAT} & \rotatebox{90}{Resisc45} & \rotatebox{90}{Retinopathy} & \rotatebox{90}{Clevr-Count} & \rotatebox{90}{Clevr-Dist} & \rotatebox{90}{DMLab} & \rotatebox{90}{KITTI-Dist} & \rotatebox{90}{dSpr-Loc} & \rotatebox{90}{dSpr-Ori} & \rotatebox{90}{sNORB-Azim} & \rotatebox{90}{sNORB-Ele} & \rotatebox{90}{\textbf{Average}}\\
\midrule
\rowstyle{\color{dt}}\multirow{1}{*}{Full~\cite{vpt}} & \rowstyle{\color{dt}}85.8 & \rowstyle{\color{dt}}68.9 & \rowstyle{\color{dt}}87.7 & \rowstyle{\color{dt}}64.3 & \rowstyle{\color{dt}}97.2 & \rowstyle{\color{dt}}86.9 & \rowstyle{\color{dt}}87.4 & \rowstyle{\color{dt}}38.8 & \rowstyle{\color{dt}}79.7 & \rowstyle{\color{dt}}95.7 & \rowstyle{\color{dt}}84.2 & \rowstyle{\color{dt}}73.9 & \rowstyle{\color{dt}}56.3 & \rowstyle{\color{dt}}58.6 & \rowstyle{\color{dt}}41.7 & \rowstyle{\color{dt}}65.5 & \rowstyle{\color{dt}}57.5 & \rowstyle{\color{dt}}46.7 & \rowstyle{\color{dt}}25.7 & \rowstyle{\color{dt}}29.1 & \rowstyle{\color{dt}}65.6\\
\rowstyle{\color{dt}}\multirow{1}{*}{Linear$^{\star}$} & \rowstyle{\color{dt}}0.04 & \rowstyle{\color{dt}}61.5 & \rowstyle{\color{dt}}88.4 & \rowstyle{\color{dt}}73.9 & \rowstyle{\color{dt}}97.9 & \rowstyle{\color{dt}}86.8 & \rowstyle{\color{dt}}41.8 & \rowstyle{\color{dt}}51.0 & \rowstyle{\color{dt}} 80.7 & \rowstyle{\color{dt}}88.6 & \rowstyle{\color{dt}}76.1 & \rowstyle{\color{dt}}74.1 & \rowstyle{\color{dt}}35.4&\rowstyle{\color{dt}}30.3&\rowstyle{\color{dt}}35.7&\rowstyle{\color{dt}}59.8&\rowstyle{\color{dt}}16.4&\rowstyle{\color{dt}}24.3&\rowstyle{\color{dt}}18.0&\rowstyle{\color{dt}}22.6&\rowstyle{\color{dt}}56.0\\
\multirow{1}{*}{Bias~\cite{vpt}} & 0.14 & 72.8&87.0&59.2&97.5&85.3&59.9&51.4&78.7&91.6&72.9&69.8&61.5&55.6&32.4&55.9&66.6&40.0&15.7&25.1&62.1\\
\multirow{1}{*}{VPT-Shallow~\cite{vpt}} & 0.11 & 77.7&86.9&62.6&97.5&87.3&74.5&51.2&78.2&92.0&75.6&72.9&50.5&58.6&40.5&67.1&68.7&36.1&20.2&34.1&64.9 \\
\multirow{1}{*}{VPT-Deep~\cite{vpt}} & 0.60 & \colorbox{lightgreen}{\textbf{78.8}}&90.8&65.8&98.0&88.3&78.1&49.6&81.8&\colorbox{lightgreen}{\textbf{96.1}}&83.4&68.4&68.5&60.0&46.5&72.8&73.6&47.9&\colorbox{lightgreen}{\textbf{32.9}}&37.8&69.4 \\
\multirow{1}{*}{Adapter~\cite{adapter}} & 0.27 &69.2&90.0&68.0&98.8&89.9&82.8&\underline{54.3}&84.0&94.9&81.9&75.5&\underline{80.9}&\underline{65.3}&48.6&78.3&74.8&48.5&29.9&\underline{41.6}&71.4\\
\multirow{1}{*}{SSF~\cite{SSF}} & 0.24 & 69.0&\colorbox{lightgreen}{\textbf{92.6}}&\colorbox{lightgreen}{\textbf{75.1}}&\colorbox{lightgreen}{\textbf{99.4}}&\colorbox{lightgreen}{\textbf{91.8}}&\underline{90.2}&52.9&\underline{87.4}&\underline{95.9}&\colorbox{lightgreen}{\textbf{87.4}}&\underline{75.5}&75.9&62.3&\underline{53.3}&\underline{80.6}&\underline{77.3}&\underline{54.9}&29.5&37.5 &\underline{73.1}\\
\midrule
\multirow{1}{*}{TTC-Tuning (ours)} & \colorbox{lightgreen}{\textbf{0.19}} &\underline{78.4}&\underline{92.4}&\underline{74.0}&\colorbox{lightgreen}{\textbf{99.4}}&\underline{91.6}&\colorbox{lightgreen}{\textbf{91.6}}&\colorbox{lightgreen}{\textbf{56.0}}&\colorbox{lightgreen}{\textbf{88.3}}&94.6&\colorbox{lightgreen}{\textbf{87.4}}&\colorbox{lightgreen}{\textbf{76.5}}&\colorbox{lightgreen}{\textbf{82.0}}&\colorbox{lightgreen}{\textbf{65.5}}&\colorbox{lightgreen}{\textbf{54.3}}&\colorbox{lightgreen}{\textbf{82.3}}&\colorbox{lightgreen}{\textbf{82.2}}&\colorbox{lightgreen}{\textbf{55.4}}&\underline{30.9}&\underline{39.1}&\colorbox{lightgreen}{\textbf{74.8}}\\
\end{tabular}
}
\end{center}
\vspace{-.2in}
\caption{\textbf{Comparisons with state-of-the-art PETL methods on the VTAB-1K benchmark with ViT-B/16}. ``$^{\star}$" means the model has been retrained to produce better results. The entries noted by {\color{dt}grey} represents the baseline algorithms. The best and second-best results of PETL methods are noted by \colorbox{lightgreen}{\textbf{green}} and \underline{underline}, respectively.}
\label{tab:vit-vtab-full}
\end{table*}

\begin{table*}[h]
\begin{center}
\large
\resizebox{0.92\linewidth}{!}{
\begin{tabular}{lc|ccccccc|cccc|cccccccc|c}
\multirow{2}{*}{} &  & \multicolumn{7}{c}{\textbf{Natural}} & \multicolumn{4}{c}{\textbf{Specialized}} & \multicolumn{8}{c}{\textbf{Structured}} \\
& \rotatebox{90}{\# Params (M)} & \rotatebox{90}{Cifar100} & \rotatebox{90}{Caltech101}  & \rotatebox{90}{DTD} & \rotatebox{90}{Flower102} & \rotatebox{90}{Pets} & \rotatebox{90}{SVHN} & \rotatebox{90}{Sun397} & \rotatebox{90}{Camelyon} & \rotatebox{90}{EuroSAT} & \rotatebox{90}{Resisc45} & \rotatebox{90}{Retinopathy} & \rotatebox{90}{Clevr-Count} & \rotatebox{90}{Clevr-Dist} & \rotatebox{90}{DMLab} & \rotatebox{90}{KITTI-Dist} & \rotatebox{90}{dSpr-Loc} & \rotatebox{90}{dSpr-Ori} & \rotatebox{90}{sNORB-Azim} & \rotatebox{90}{sNORB-Ele} & \rotatebox{90}{\textbf{Average}}\\
\midrule

 \rowstyle{\color{dt}} Full~\cite{vpt} &  \rowstyle{\color{dt}} 86.7 & \rowstyle{\color{dt}}72.2& \rowstyle{\color{dt}}88.0& \rowstyle{\color{dt}}71.4& \rowstyle{\color{dt}}98.3& \rowstyle{\color{dt}}89.5& \rowstyle{\color{dt}}90.1& \rowstyle{\color{dt}}45.0& \rowstyle{\color{dt}}86.6& \rowstyle{\color{dt}}96.9& \rowstyle{\color{dt}}87.7& \rowstyle{\color{dt}}79.4& \rowstyle{\color{dt}}75.7& \rowstyle{\color{dt}}59.8& \rowstyle{\color{dt}}54.6& \rowstyle{\color{dt}}78.6& \rowstyle{\color{dt}}79.4& \rowstyle{\color{dt}}53.6& \rowstyle{\color{dt}}34.6& \rowstyle{\color{dt}}40.9&\rowstyle{\color{dt}}74.2\\

\rowstyle{\color{dt}} Linear~\cite{vpt} & \rowstyle{\color{dt}}0.04&\rowstyle{\color{dt}}61.4&\rowstyle{\color{dt}}90.2&\rowstyle{\color{dt}}74.8&\rowstyle{\color{dt}}99.5&\rowstyle{\color{dt}}90.2&\rowstyle{\color{dt}}42.7&\rowstyle{\color{dt}}55.8&\rowstyle{\color{dt}}81.5&\rowstyle{\color{dt}}90.1&\rowstyle{\color{dt}}82.1&\rowstyle{\color{dt}}69.4&\rowstyle{\color{dt}}39.1&\rowstyle{\color{dt}}35.9&\rowstyle{\color{dt}}40.1&\rowstyle{\color{dt}}65.0&\rowstyle{\color{dt}}20.3&\rowstyle{\color{dt}}26.0&\rowstyle{\color{dt}}14.3&\rowstyle{\color{dt}}27.8&\rowstyle{\color{dt}}56.4\\
\multirow{1}{*}{Bias~\cite{vpt}} & 0.29 &73.0&86.8&65.6& \colorbox{lightgreen}{{\textbf{97.7}}}&87.5&\underline{56.4}&\underline{52.3}&80.4&91.6&76.1&\underline{72.5}&47.3&48.5&34.7&\underline{66.2}&57.6&36.2& \colorbox{lightgreen}{{\textbf{34.7}}}& \colorbox{lightgreen}{{\textbf{66.2}}}&62.1\\

\multirow{1}{*}{VPT-Deep~\cite{vpt}} & 0.24 &\colorbox{lightgreen}{{\textbf{79.6}}}&\underline{90.8}&\colorbox{lightgreen}{{\textbf{78.0}}}&99.5&\underline{91.4}&42.3&51.7&\underline{84.9}& \colorbox{lightgreen}{{\textbf{96.2}}}&\underline{85.0}&72.0&\underline{67.6}&\underline{59.4}&\underline{50.1}&61.3&\underline{74.4}&\underline{50.6}&25.7&25.7&\underline{68.6}\\
\multirow{1}{*}{TTC-Tuning (ours)} & 0.19 & \underline{76.1} & \colorbox{lightgreen}{{\textbf{92.4}}} & \underline{76.6}& \colorbox{lightgreen}{{\textbf{99.7}}}& \colorbox{lightgreen}{{\textbf{92.8}}} &  \colorbox{lightgreen}{{\textbf{88.5}}} &  \colorbox{lightgreen}{{\textbf{55.1}}}& \colorbox{lightgreen}{{\textbf{88.0}}}& \underline{95.8}& \colorbox{lightgreen}{{\textbf{87.5}}}& \colorbox{lightgreen}{{\textbf{75.4}}}& \colorbox{lightgreen}{{\textbf{82.3}}} & \colorbox{lightgreen}{{\textbf{62.5}}} &  \colorbox{lightgreen}{{\textbf{52.4}}} & \colorbox{lightgreen}{{\textbf{83.4}}} & \colorbox{lightgreen}{{\textbf{82.6}}}& \colorbox{lightgreen}{{\textbf{54.3}}} & \underline{30.6} & \underline{39.8}& \colorbox{lightgreen}{{\textbf{74.5}}}\\
\end{tabular}
}
\end{center}
\vspace{-.2in}
\caption{\textbf{Comparisons with state-of-the-art methods on the VTAB-1K benchmark with Swin-B}.}
\label{tab:swin-vtab-full}
\vspace{-.2in}
\end{table*}

\begin{table*}[t]
\vspace{-0.2in}
\begin{center}
\large
\resizebox{0.92\linewidth}{!}{
\begin{tabular}{c|cccc|ccccccc|cccc|cccccccc|c}
& &&&  & \multicolumn{7}{c}{\textbf{Natural}} & \multicolumn{4}{c}{\textbf{Specialized}} & \multicolumn{8}{c}{\textbf{Structured}} & \\
No. &\rotatebox{90}{LN Tuning Stage1} & \rotatebox{90}{LN Tuning Stage2} & \rotatebox{90}{TTC} & \rotatebox{90}{\# Params (M)} & \rotatebox{90}{Cifar100} & \rotatebox{90}{Caltech101}  & \rotatebox{90}{DTD} & \rotatebox{90}{Flower102} & \rotatebox{90}{Pets} & \rotatebox{90}{SVHN} & \rotatebox{90}{Sun397} & \rotatebox{90}{Camelyon} & \rotatebox{90}{EuroSAT} & \rotatebox{90}{Resisc45} & \rotatebox{90}{Retinopathy} & \rotatebox{90}{Clevr-Count} & \rotatebox{90}{Clevr-Dist} & \rotatebox{90}{DMLab} & \rotatebox{90}{KITTI-Dist} & \rotatebox{90}{dSpr-Loc} & \rotatebox{90}{dSpr-Ori} & \rotatebox{90}{sNORB-Azim} & \rotatebox{90}{sNORB-Ele} & \rotatebox{90}{\textbf{Average}}\\
\midrule

1&\xmark & \xmark & \xmark & 0.04 & 61.5 & 88.4 & 73.9 & 97.9 & 86.8 & 41.8 & 51.0 &  80.7 & 88.6 & 76.1 & 74.1 &35.4&30.3&35.7&59.8&16.4&24.3&18.0&22.6&56.0\\
2&\cmark & \xmark & \xmark & 0.08 &74.9&91.6&\colorbox{lightgreen}{{\textbf{75.2}}}&99.2&91.4&90.5&55.5&86.6&\colorbox{lightgreen}{{\textbf{95.9}}}&87.1&76.1&80.6&65.0&53.1&80.9&75.5&55.4&25.5&37.5&73.6\\


3&\xmark & \xmark & \cmark & 0.15& 74.7& 91.6&73.6&99.1&90.8&90.2&52.9&87.4&95.4&86.4&75.1&80.5&63.8&51.5&80.6&78.9&55.5&28.6&\colorbox{lightgreen}{{\textbf{40.2}}}&73.1\\

4&\cmark & \xmark & \cmark & 0.15&77.3&91.7&72.9&99.4&91.1&90.6&54.4&84.2&94.3&87.3&75.4&\colorbox{lightgreen}{{\textbf{82.0}}}&65.1&53.0&80.9&82.1&\colorbox{lightgreen}{{\textbf{55.9}}}&28.8&38.2&73.9\\

5&\xmark & \cmark & \cmark & 0.19 & 75.0&91.9&73.9&99.4&90.8&90.8&54.6&87.6&95.8&\colorbox{lightgreen}{{\textbf{87.6}}}&75.3&80.8&64.0&52.4&80.5&82.1&55.3&29.0&40.1&74.1\\

6&\cmark &\cmark & \cmark & 0.19 &\colorbox{lightgreen}{{\textbf{78.4}}}&
\colorbox{lightgreen}{{\textbf{92.4}}}&74.0&\colorbox{lightgreen}{{\textbf{99.4}}}&\colorbox{lightgreen}{{\textbf{91.6}}}&\colorbox{lightgreen}{{\textbf{91.6}}}&\colorbox{lightgreen}{{\textbf{56.0}}}&\colorbox{lightgreen}{{\textbf{88.3}}}&94.6&87.4&\colorbox{lightgreen}{{\textbf{76.5}}}&\colorbox{lightgreen}{{\textbf{82.0}}}&\colorbox{lightgreen}{{\textbf{65.5}}}&\colorbox{lightgreen}{{\textbf{54.3}}}&\colorbox{lightgreen}{{\textbf{82.3}}}&\colorbox{lightgreen}{{\textbf{82.2}}}&55.4&\colorbox{lightgreen}{{\textbf{30.9}}}&39.1&\colorbox{lightgreen}{{\textbf{74.8}}}\\
\end{tabular}
}
\end{center}
\vspace{-.2in}
\caption{\textbf{Evaluation of our proposed Paradigm}.}
\vspace{-.2in}
\label{tab:vit-vtab-full-evalmethod}
\end{table*}

\section{Experiments}


\subsection{Experiments on VTAB-1K Benchmark}

\noindent\textbf{Dataset.}
VTAB-1K~\cite{zhai2019large} contains 19 visual classification tasks which cover a broad spectrum of domains and semantics in three groups, \ie, \textit{Natural}, \textit{Specialized}, and \textit{Structured}. The \textit{Natural} group contains 7 classic classification datasets~\cite{krizhevsky2009learning, fei2004learning, cimpoi14describing, nilsback2006visual, parkhi2012cats, netzer2011reading, xiao2010sun} of natural images. The \textit{Specialized} group involves 4 datasets~\cite{Veeling2018qh, helber2019eurosat, cheng2017remote, kaggle2015retinopathy} of two special scenarios: medical and remote-sensing. The \textit{Structured} group has 8 datasets~\cite{johnson2017clevr, beattie2016deepmind, geiger2013vision, matthey2017dsprites, lecun2004learning}, mainly focusing on understanding the structure of a scene, such as object counting, and depth prediction. Each task of VTAB-1K contains 1000 training images.
Following \cite{vpt, SSF}, we use the 800-200 \textsc{train-val} split to determine the hyperparameters and the entire 1000 training data to train the final model. 
We report the average top-1 accuracy on the \textsc{test} set. 

\noindent\textbf{Baselines and state-of-the-art approaches.}
We compare our method with three baselines, Full fine-tuning, Linear, and Bias, and three state-of-the-art methods Adapter~\cite{adapter}, VPT~\cite{vpt}, and SSF~\cite{SSF}. Bias method only updates all the bias terms in the pre-trained backbone. 

\noindent\textbf{Performance with ViT backbone.}
We compare our TTC-tuning with the above 7 baselines in Tab.~\ref{tab:vit-vtab-full}. We use ViT-B/16 as the backbone and insert TTC-Module in each transformer layer. The default $K$ is set to 96, 1/8 of the total channels, leading to the trainable parameter number being only 0.11M.
\textbf{First}, our TTC-Tuning achieves the average accuracy of 74.8\% on the 19 downstream tasks, outperforming the full fine-tuning on 18 out of 19 tasks and gains the improvement of 6.2\%, 3.3\%, and 13.9\% in the three groups, respectively, with only additional 0.13\% of the backbone parameters. Such results reflect that TTC-Tuning can greatly reduce the storage space and alleviate the overfitting problem commonly occurring in full fine-tuning large models.
\textbf{Second}, compared with Adapter~\cite{adapter} that treats all the channels equally, selecting a part of task-relevant channels for each downstream task is more effective and efficient, outperforming it by 3.4\% in average accuracy. Moreover, our TTC-Tuning outperforms VPT~\cite{vpt} by 5.0\%, 4.3\%, and 6.5\% in the three groups, respectively.
\textbf{Third}, compared with the distribution alignment method SSF~\cite{SSF}, our TTC-Tuning surpasses it by 1.7\%. 
These results demonstrate that instead of aligning distribution only (\ie, SSF) or learning task-relevant information (\ie, VPT, Adapter), leveraging the two-stage paradigm can maintain lower-level parameter costs and improve the performance.

\noindent\textbf{Performance with Swin Transformer Backbone.}
To verify the effectiveness of TTC-Tuning with different backbones, we apply TTC-Tuning on hierarchical transformers, \ie, Swin-B~\cite{liu2021swin}. We use the same setting of inserting TTC-Module as in the ViT backbone.
Considering deep layers contain more semantic information in the hierarchical structure, instead of applying TTC-Module on all the transformer layers, we insert it to the last half of the layers in the stage3 and all layers of the stage4 of the Swin-B to keep a similar level of trainable parameters.
The results of Tab. \ref{tab:swin-vtab-full} show that TTC-Tuning outperforms \textbf{Full fine-tuning} in all three groups with only 0.2\% parameters while other methods cannot. In addition, compared with PETL method, TTC-Tuning outperforms VPT~\cite{vpt} by 6.2\%, 2.2\%, and 7.6\% in the three groups, respectively. 
All the results above suggest that our TTC-tuning is also applicable for the hierarchical transformers and can yield much more improvement than other PETL methods.

\noindent\textbf{Complexity Analysis.} In our analysis, we consider a ViT-B backbone with $L$ layers and $D$ dimensions, along with $N$ tokens for a single image. We also assume that the intermediate dimension of Adapter~\cite{adapter} is $D^{\prime}$, that the prompt length of VPT~\cite{vpt} is $n$, and that the total insert times of SSF~\cite{SSF} is $m$ in the whole ViT-B backbone. Finally, we compare our proposed TTC-Module approach to Adapter, VPT, and SSF in terms of parameters and FLOPs, as summarized in Tab.~\ref{tab:complexity}. Notably, our selection of $K$ as $\frac{1}{8}D$ is fairly small compared to $D$. When we compare our approach to SSF, we find that the number of parameters for TTC-Module is $\frac{1}{64}LDD$, while the number of parameters for SSF is $mLD$. Examining the ViT-B backbone, we find that $m=74$ and $\frac{1}{64}D=12$, our parameters and FLOPs are smaller than SSF. Overall, our analysis suggests that TTC-Module may offer a more efficient and effective approach to transfer learning.

\begin{table}[t]
\small
\centering
\resizebox{0.9\linewidth}{!}{
\begin{tabular}{l|cccc}
\toprule
& Adapter & VPT-Deep & SSF  & TTC-Module \\
\midrule
\multirow{1}{*}{\# Extra Parameters} & $2LDD^{\prime}$& $nLD$ & $mLD$ & $LKK$ \\
\multirow{1}{*}{\# Extra FLOPs} & $2NLDD^{\prime}$ & $2n(2N+n)LD$ & $mNLD$ & $NLKK$  \\
\bottomrule
\end{tabular}
}
\caption{A complexity analysis of Adapter~\cite{adapter}, VPT~\cite{vpt}, SSF~\cite{SSF}, and our proposed TTC-Module.}
\label{tab:complexity}
\vspace{-0.2in}
\end{table}

\subsection{Evaluation}

\begin{table*}[th]
\vspace{-.3in}
\centering
\subfloat[
\textbf{Channel Selection}.
\label{tab:ablation:differentSS}
]{
\begin{minipage}{0.18\linewidth}{\begin{center}
\scriptsize
\setlength{\tabcolsep}{1.2mm}{
\begin{tabular}{c|cc}
 & Acc. & Params. (M) \\
\midrule
\multirow{1}{*}{Linear$^{\star}$} & 61.5 & 0.08 \\
\multirow{1}{*}{RC-1} & 72.1 & 0.23 \\
\multirow{1}{*}{RC-2} &74.1& 0.23  \\
\multirow{1}{*}{RC-3} &71.3& 0.23  \\
\multirow{1}{*}{L2 Norm} &75.4 & 0.23  \\
\multirow{1}{*}{TIS} & \colorbox{lightgreen}{{\textbf{78.4}}}& 0.23 \\
\end{tabular}}
\end{center}}\end{minipage}}
\hspace{2em}
\subfloat[
\textbf{Insert depth}.
\label{tab:ablation:insertdepth}
]{
\begin{minipage}{0.16\linewidth}{
\begin{center}
\scriptsize
\setlength{\tabcolsep}{1.5mm}{
\begin{tabular}{c|cc}
\multirow{1}{*}{Layers } & Acc. & Params. (M) \\
\midrule
\multirow{1}{*}{Linear$^{\star}$} & 61.5 & 0.08 \\
\multirow{1}{*}{0} & 74.9 & 0.12 \\
\multirow{1}{*}{2} & 77.8 & 0.13 \\
\multirow{1}{*}{4} & 77.7 &  0.15\\
\multirow{1}{*}{8} & 77.9 &  0.19\\
\multirow{1}{*}{12} & \colorbox{lightgreen}{{\textbf{78.4}}} & 0.23\\
\end{tabular}}
\end{center}}\end{minipage}
}
\hspace{2em}
\subfloat[
\textbf{Insert position}. 
\label{tab:ablation:insetrposition}
]{
\begin{minipage}{0.22\linewidth}{\begin{center}
\scriptsize
\setlength{\tabcolsep}{1.5mm}{
\begin{tabular}{c|cc}
\multirow{1}{*}{Insert Position} & Acc. & \# Params. (M) \\
\midrule
\multirow{1}{*}{Full} & 68.9& 86.7\\
\multirow{1}{*}{Linear$^{\star}$} & 61.5 & 0.08 \\
\multirow{1}{*}{LayerNorm$^{\star}$} & 73.6 & 0.11 \\
\multirow{1}{*}{MHSA} & 78.4 &  \multirow{1}{*}{0.23}\\
\multirow{1}{*}{MLP} & \colorbox{lightgreen}{{\textbf{78.6}}} & \multirow{1}{*}{0.23}\\
\multirow{1}{*}{MHSA+MLP} & 77.1 &  \multirow{1}{*}{0.34} \\
\end{tabular}}
\end{center}}\end{minipage}
}
\hspace{2em}
\subfloat[
\textbf{Different $K$}.
\label{tab:ablation:differntK}
]{
\begin{minipage}{0.18\linewidth}{\begin{center}
\scriptsize
\setlength{\tabcolsep}{1.5mm}{
\begin{tabular}{c|cc}
\multirow{1}{*}{Top-$K$ } & Acc. & Params. (M) \\
\midrule
\multirow{1}{*}{Linear$^{\star}$} & 61.5 & 0.08 \\
\multirow{1}{*}{32} & 78.2 & 0.13 \\
\multirow{1}{*}{64} & 77.7 & 0.17\\
\multirow{1}{*}{96} & \colorbox{lightgreen}{{\textbf{78.4}}} &  0.23 \\
\multirow{1}{*}{128} & 77.4 & 0.31 \\
\multirow{1}{*}{192} & 76.5 & 0.56\\
\end{tabular}
}
\end{center}}\end{minipage}
}
\vspace{-.1in}
\caption{\textbf{Evaluation of different designs}. Acc.: Top-1 accuracy (\%); Params.: parameters (M). \text{Linear$^{\star}$} represents the baseline results for better comparison. }

 
\label{tab:ablations}
\vspace{-.1in}
\end{table*}

\begin{figure*}[t]
\begin{center}
\includegraphics[width=0.23\linewidth]{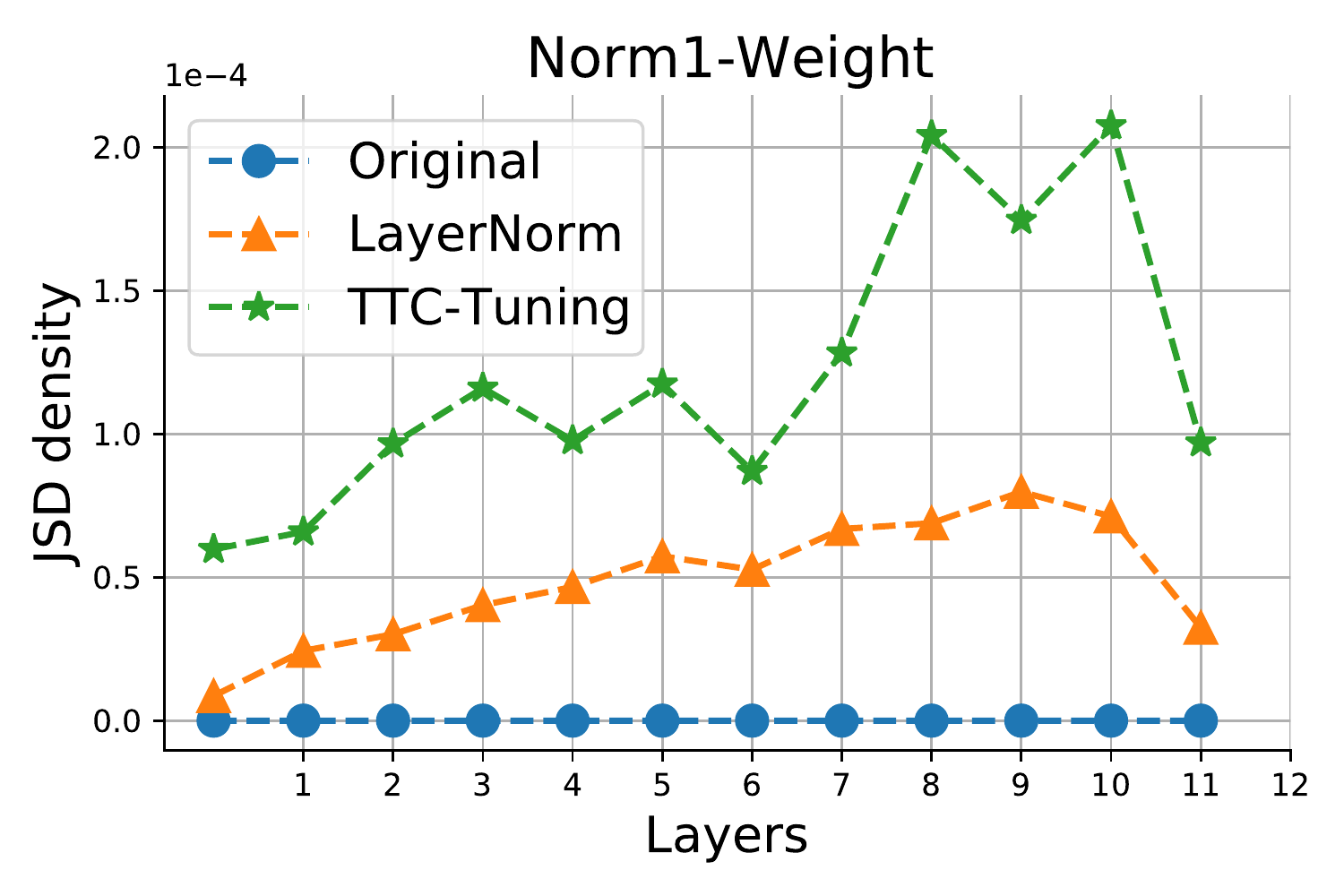}
\includegraphics[width=0.23\linewidth]{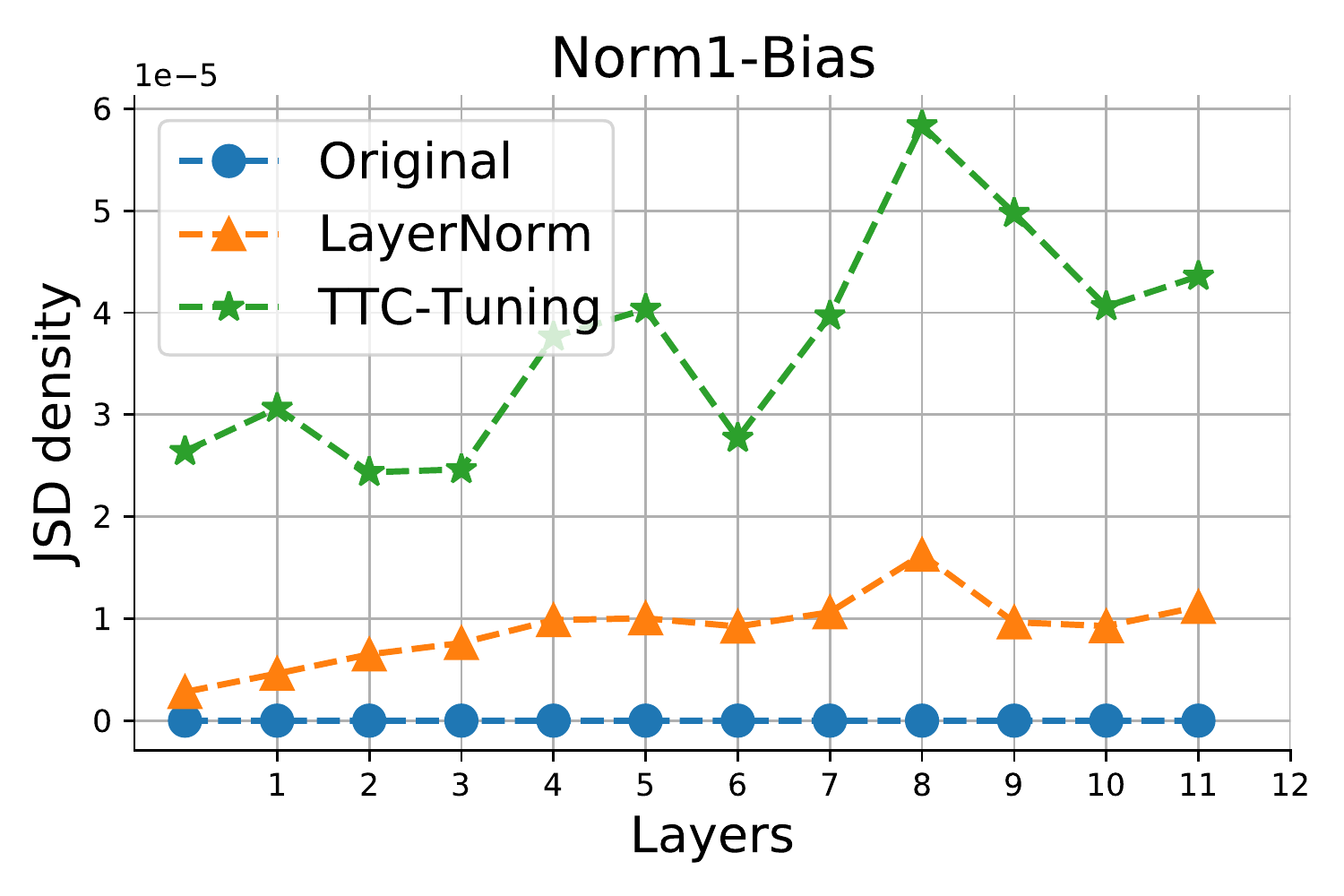}
\includegraphics[width=0.23\linewidth]{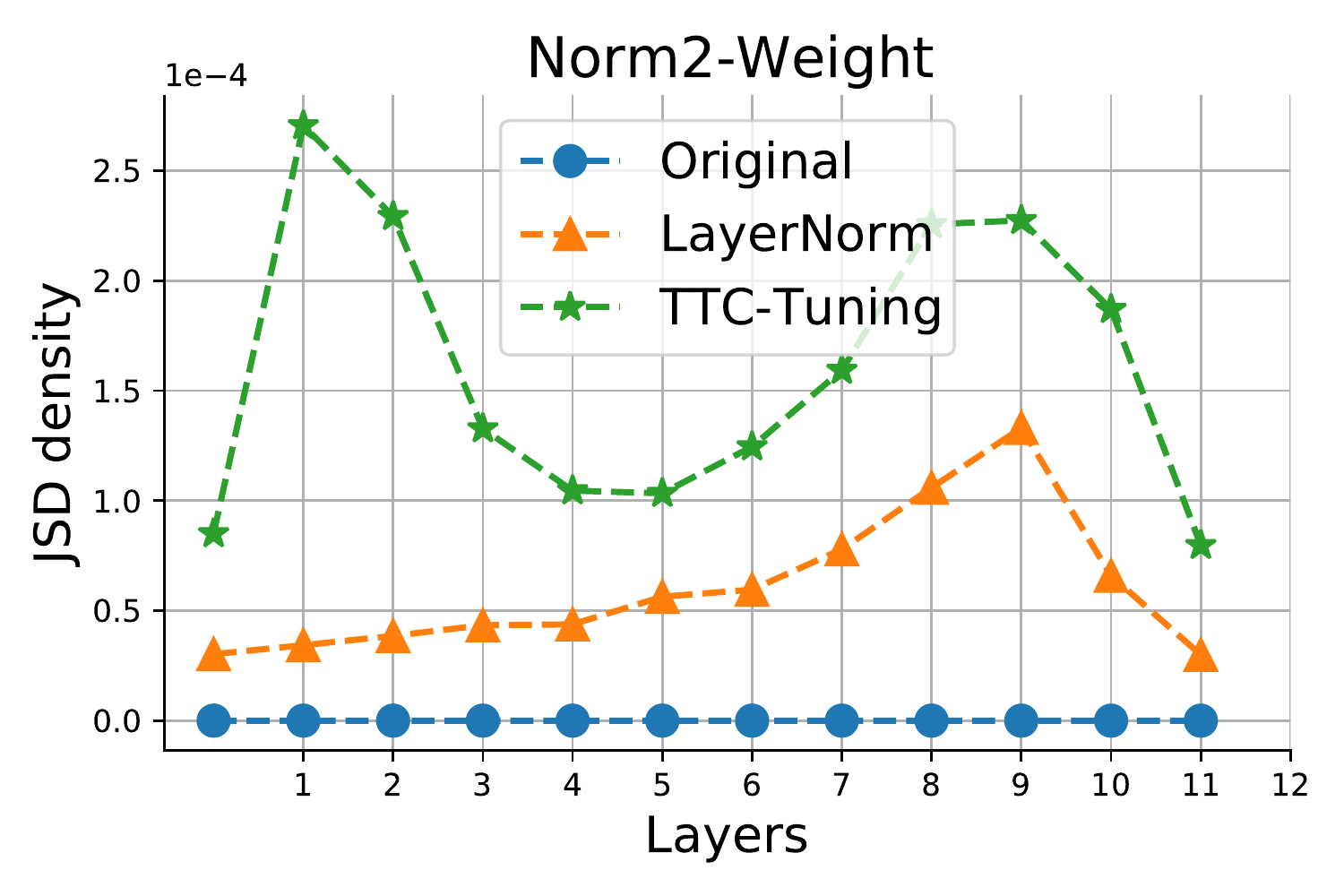}
\includegraphics[width=0.23\linewidth]{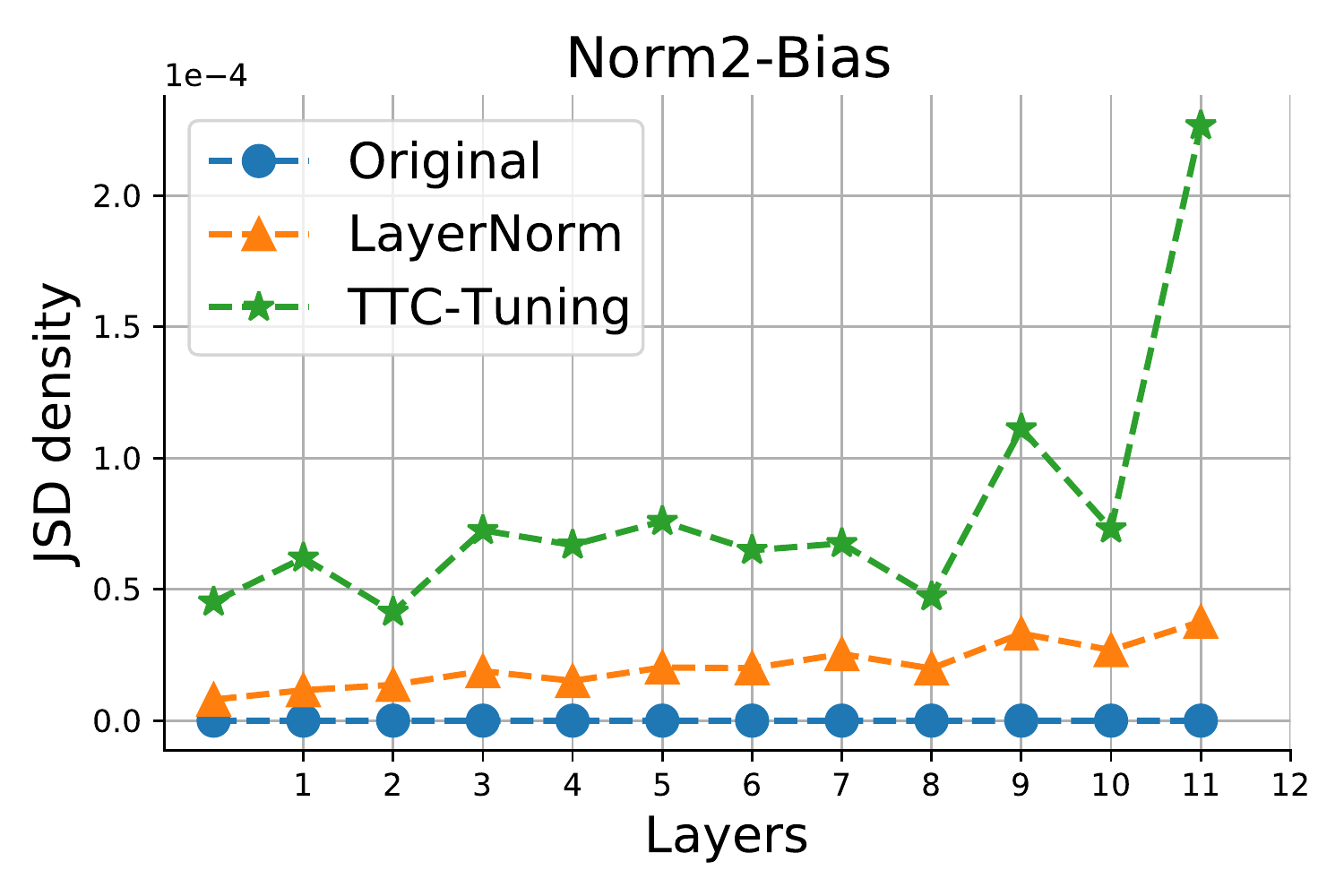}
\end{center}
 \vspace{-0.3in}
\caption{Comparison of parameter shift after tuning the LayerNorm layer (stage1) and jointly tuning LayerNorm and TTC-Module (stage2).}
\label{fig:scalebiaschange}
\vspace{-0.2in}
\end{figure*}

\noindent\textbf{Ablation Studies.}
To evaluate the effectiveness of tuning the LayerNorm layer and our proposed TTC-Module, we conduct ablation studies on the two components  in our two-stage paradigm in Tab.~\ref{tab:vit-vtab-full-evalmethod}.
\textbf{First}, in the first stage, we finetune LayerNorm to align task distribution, which has already outperformed the previous best model (73.6\% (2nd row of Tab.~\ref{tab:vit-vtab-full-evalmethod}) \textit{v.s.} 73.1\% (SSF). \textbf{Second}, when combining the second stage on top of the LayerNorm tuning, the TTC-Module can yield an improvement of 1.2\%.
\textbf{Third}, to further verify the effectiveness of TTC-Module, we insert TTC directly into the baseline \textit{Linear} model, gaining an improvement of 17.1\% (4th row). 
\textbf{Fourth}, we tune the LayerNorm and TTC-Module together in one stage (5th row), achieving an accuracy of 74.1\%, worse than the two-stage paradigm by 0.7\%.
All the results above demonstrate the effectiveness of the proposed LayerNorm tuning, TTC-Module, and the necessity of a two-stage paradigm.

\noindent\textbf{Tuning Channels \vs Tuning Weights}
As illustrated in Sec.~\ref{sec:adapt}, tuning selected $K$ channels via a linear adapter only use $\frac{K}{D}=\frac{1}{8}$ parameters of directly tuning the weights of ViT layer. In addition, with fewer learnable parameters, the model is less prone to overfitting to the small dataset.
We compare the performance of tuning weights and tuning channels in Tab.~\ref{tab:vit-vtab-weights-channels}. The number of parameters of tuning weights is relatively high (0.88M) while tuning channels with only 0.11M parameters can gain 7.8\% improvements in total 19 downstream tasks.

\noindent\textbf{Effectiveness of Task-Relevant Channel Selection.}
To verify the effectiveness and necessity of the proposed Taylor expansion-based Importance Score (TIS) channel selection, we compare three channel selection strategies in Tab.~\ref{tab:ablation:differentSS}. These strategies included Random Channel Selection (RC), L2 Norm, and Taylor expansion-based Importance Score (TIS). RC selects $K$ channels randomly, and to reduce the impact of outliers, we randomly selected three sets of channels (RC-1/2/3). L2 Norm determines task-relevant channels based on the L2 Norm of features in each channel. The task-relevant strategies achieve better and more robust performance than RC. In addition, our TIS can select more important and representative channels than L2 Norm, outperforming it by 3.0\%. 

\noindent\textbf{Insert Depth.}
Insert depth is one important factor that influences performance. We report the results when inserting TTC-Module to the last $l$ layers of ViT-B in Tab.~\ref{tab:ablation:insertdepth}. Without the TTC-Module (stage one only), the accuracy is 74.9\%, while the accuracy gradually improved to 78.4\% with the increase of insert depth. 
Upon analyzing the results in Tab.~\ref{tab:ablation:insertdepth}, we observe that inserting the TTC-Module only in the last two layers achieved an accuracy of 77.8\%, indicating that deeper layers contribute more to the final results. Notably, when we remove the TTC-Module in the first four layers, the accuracy was 77.9\%, with only a 0.5\% gap to the best result of 78.4\%.

\noindent\textbf{Insert Position.}
We evaluated the insertion position of our TTC-Module, as shown in Tab.~\ref{tab:ablation:insetrposition}. Specifically, we insert the module after the MHSA and MLP blocks, respectively. Our findings indicate that inserting the module after the MLP block yields better results, consistent with similar findings for SSF. Additionally, we insert our module after both blocks, which led to lower performance at 77.1\% with an increase in parameters. We conjecture that only one position is enough for adaptation, and repeat adaptation will increase the difficulty of optimization.

\noindent\textbf{Number of selected channels $K$.}
The number of selected channels ($K$) is the most important hyperparameter related to the design of TTC-Module, influencing the model architecture and the number of trainable parameters.
Unlike VPT, which selects the best prompt length for each task, we use the same $K$ for all tasks for a fair comparison. In Tab.~\ref{tab:ablation:differntK}, as we increase the value of $K$, the performance improves and peaks at $K=96$.  When further increasing the learnable channels, the performance degrades. 
We hypothesize that a larger value of $K$ may involve too much task-irrelevant information and can make tuning hyperparameters more difficult.

\noindent\textbf{Analysis.}
In Figure~\ref{fig:scalebiaschange}, we analyzed the parameter shift after tuning the LayerNorm layer (stage1) and jointly tuning LayerNorm and TTC-Module (stage2). Our findings indicate that deeper layers led to larger shifts in the weight and bias of both ``Norm1" and ``Norm2" (the two LayerNorm layers in ViT-B). Specifically, we observed obvious deviations in the weight parameter of the shallow layers of "Norm2" which differed from "Norm1". We also evaluate the representation ability to conduct stage1 tuning in Tab.~\ref{tab:} by utilizing KNN~\cite{cover1967nearest} algorithm to cluster the feature of \textsc{[CLS]} token. The results suggest that stage1 is indeed effective in improving representation ability.

\subsection{Experiments on Domain Generalization}
In addition to evaluating the model on test data of the same distribution, modern deep neural networks commonly suffer from performance degradation when the testing distribution is different from that of the training set, \ie, domain shift, which is inevitable in a real-world application. 
\noindent\textbf{Dataset.}
We use the ImageNet-1K~\cite{deng2009imagenet} as the source domain with 16-shot per category and evaluate our model on ImageNetV2~\cite{recht2019imagenet}, ImageNet-Sketch~\cite{wang2019learning}, ImageNet-A~\cite{hendrycks2021natural}, and ImageNet-R~\cite{hendrycks2021many}.


\noindent\textbf{Results.}
In Tab.~\ref{tab:dg}, we compare our TTC-tuning with Adapter~\cite{adapter}, VPT~\cite{vpt}, LoRA~\cite{lora}, and NOAH~\cite{noah} on the above datasets. We can make two observations. \textbf{First}, TTC-tuning outperforms the previous best method~(NOAH) on three of the four target datasets and achieves comparable performance on ImageNetV2. Specifically, TTC-tuning yields an improvement of 0.9\% on ImageNet-R over NOAH. \textbf{Second}, our TTC-tuning achieves an accuracy of 75.5\% on the source domain, greatly outperforming previous methods by 4\%. 
%
Since the backbone model is pre-trained on ImageNet-21K, the results on ImageNet-1K show that TTC-tuning can better align the superset's complex distribution with the subset's relatively simple distribution. The two observations demonstrate the superiority of our TTC-tuning over previous PETL techniques on strong generalization ability.

\begin{table}[]
\small
\centering
\resizebox{0.3\textheight}{!}{
\begin{tabular}{l|c|cccc}
\toprule
\multirow{2}{4em}{} & {\textbf{Source}} & \multicolumn{4}{c}{\textbf{Target}} \\
& ImageNet & -V2 & -Sketch  & -A & -R \\
\midrule
\multirow{1}{*}{Adapter~\cite{adapter}}  & 70.5 & 59.1 & 16.4 & 5.5 & 22.1 \\
\multirow{1}{*}{VPT~\cite{vpt}}  & 70.5 & 58.0 & 18.3 & 4.6 & 23.2 \\
\multirow{1}{*}{LoRA~\cite{lora}}  & 70.8 & 59.3 & 20.0 & 6.9 & 23.3 \\
\multirow{1}{*}{NOAH~\cite{noah}} & 71.5 & \colorbox{lightgreen}{{\textbf{66.1}}} & 24.8 & \colorbox{lightgreen}{{\textbf{11.9}}} & 28.5 \\
\midrule
\multirow{1}{*}{TTC-Tuning (ours)} & \colorbox{lightgreen}{{\textbf{75.5}}} & 65.9 & \colorbox{lightgreen}{{\textbf{25.6}}} & \colorbox{lightgreen}{{\textbf{11.9}}} & \colorbox{lightgreen}{{\textbf{29.4}}}\\
\bottomrule
\end{tabular}
}
\vspace{-.1in}
\caption{Comparison with previous methods on domain generalization.}
\label{tab:dg}
\vspace{-0.1in}
\end{table}

\begin{table}[]
\begin{center}
\resizebox{0.94\linewidth}{!}{
\begin{tabular}{l|c|cccc}
\toprule
\multirow{2}{4em}{} & \multirow{2}{*}{{\#} Params (M)} & \multicolumn{4}{c}{VTAB-1K} \\
& & Natural & Specialized  & Structured & Average \\

\midrule
\multirow{1}{*}{Tuning weights} & 0.88+0.08 & 78.1 & 84.2 & 57.1 & 72.9 \\
\multirow{1}{*}{Tuning Channels} & 0.11+0.08 & \colorbox{lightgreen}{{\textbf{83.4}}}$_{+5.3}$ & \colorbox{lightgreen}{{\textbf{86.7}}}$_{+2.5}$ & \colorbox{lightgreen}{{\textbf{61.5}}}$_{+4.4}$ & \colorbox{lightgreen}{{\textbf{74.8}}}$_{+1.9}$ \\

\bottomrule
\end{tabular}
}
\end{center}
\vspace{-0.2in}
\caption{We evaluated the effects of tuning task-relevant weights and channels on 19 downstream tasks. The a+b notation represents the combination of parameters introduced by both external and internal modules.}
\label{tab:vit-vtab-weights-channels}
\vspace{-0.1in}
\end{table}

\begin{table}[]
\begin{center}
\scriptsize
\begin{tabular}{l|ccc}
\toprule
Method & SVHN & EuroSAT & Clevr-Count \\
\midrule

w/o stage1 & 34.50& 84.44& 29.00\\
w/ stage1 & 91.23& 92.96&80.33\\

\bottomrule
\end{tabular}
\end{center}
\vspace{-0.2in}
\caption{We evaluate the stage1 using the KNN algorithm to test the representation ability of \textsc{[CLS]} token.}
\label{tab:}
\vspace{-0.2in}
\end{table}

\section{Conclusion}
Since previous PETL methods can be divided into two streams: learning task-relevant information and aligning the distributions between pre-trained and downstream tasks, we first propose a two-stage paradigm by combining the two lines of approaches. We first narrow the distribution shifts and propose a Taylor expansion-based importance score to select task-relevant channels for efficient adaption. In summary, our novel paradigm represents a new direction emphasizing the importance of considering distribution shifts when fine-tuning downstream tasks.

{\small
\bibliographystyle{ieee_fullname}
\bibliography{egbib}

\begin{thebibliography}{10}\itemsep=-1pt

\bibitem{ali2021xcit}
Alaaeldin Ali, Hugo Touvron, Mathilde Caron, Piotr Bojanowski, Matthijs Douze,
  Armand Joulin, Ivan Laptev, Natalia Neverova, Gabriel Synnaeve, Jakob
  Verbeek, et~al.
\newblock Xcit: Cross-covariance image transformers.
\newblock In {\em NeurIPS}, 2021.

\bibitem{ba2016layer}
Jimmy~Lei Ba, Jamie~Ryan Kiros, and Geoffrey~E Hinton.
\newblock Layer normalization.
\newblock {\em arXiv preprint arXiv:1607.06450}, 2016.

\bibitem{beattie2016deepmind}
Charles Beattie, Joel~Z Leibo, Denis Teplyashin, Tom Ward, Marcus Wainwright,
  Heinrich K{\"u}ttler, Andrew Lefrancq, Simon Green, V{\'\i}ctor Vald{\'e}s,
  Amir Sadik, et~al.
\newblock Deepmind lab.
\newblock {\em arXiv preprint arXiv:1612.03801}, 2016.

\bibitem{brown2020language}
Tom Brown, Benjamin Mann, Nick Ryder, Melanie Subbiah, Jared~D Kaplan, Prafulla
  Dhariwal, Arvind Neelakantan, Pranav Shyam, Girish Sastry, Amanda Askell,
  et~al.
\newblock Language models are few-shot learners.
\newblock In {\em NeurIPS}, 2020.

\bibitem{chen2021crossvit}
Chun-Fu~Richard Chen, Quanfu Fan, and Rameswar Panda.
\newblock Crossvit: Cross-attention multi-scale vision transformer for image
  classification.
\newblock In {\em Proceedings of the IEEE/CVF international conference on
  computer vision}, pages 357--366, 2021.

\bibitem{chen2022conv}
Hao Chen, Ran Tao, Han Zhang, Yidong Wang, Wei Ye, Jindong Wang, Guosheng Hu,
  and Marios Savvides.
\newblock Conv-adapter: Exploring parameter efficient transfer learning for
  convnets.
\newblock {\em arXiv preprint arXiv:2208.07463}, 2022.

\bibitem{chen2022adaptformer}
Shoufa Chen, Chongjian Ge, Zhan Tong, Jiangliu Wang, Yibing Song, Jue Wang, and
  Ping Luo.
\newblock Adaptformer: Adapting vision transformers for scalable visual
  recognition.
\newblock {\em arXiv preprint arXiv:2205.13535}, 2022.

\bibitem{cheng2017remote}
Gong Cheng, Junwei Han, and Xiaoqiang Lu.
\newblock Remote sensing image scene classification: Benchmark and state of the
  art.
\newblock {\em Proceedings of the IEEE}, 105(10):1865--1883, 2017.

\bibitem{cimpoi14describing}
M. Cimpoi, S. Maji, I. Kokkinos, S. Mohamed, , and A. Vedaldi.
\newblock Describing textures in the wild.
\newblock In {\em Proceedings of the {IEEE} Conf. on Computer Vision and
  Pattern Recognition ({CVPR})}, 2014.

\bibitem{cover1967nearest}
Thomas Cover and Peter Hart.
\newblock Nearest neighbor pattern classification.
\newblock {\em IEEE transactions on information theory}, 13(1):21--27, 1967.

\bibitem{deng2009imagenet}
Jia Deng, Wei Dong, Richard Socher, Li-Jia Li, Kai Li, and Li Fei-Fei.
\newblock Imagenet: A large-scale hierarchical image database.
\newblock In {\em Proceedings of the IEEE/CVF Conference on Computer Vision and
  Pattern Recognition (CVPR)}, pages 248--255. Ieee, 2009.

\bibitem{devlin2018bert}
Jacob Devlin, Ming-Wei Chang, Kenton Lee, and Kristina Toutanova.
\newblock Bert: Pre-training of deep bidirectional transformers for language
  understanding.
\newblock {\em arXiv preprint arXiv:1810.04805}, 2018.

\bibitem{dong2022cswin}
Xiaoyi Dong, Jianmin Bao, Dongdong Chen, Weiming Zhang, Nenghai Yu, Lu Yuan,
  Dong Chen, and Baining Guo.
\newblock Cswin transformer: A general vision transformer backbone with
  cross-shaped windows.
\newblock In {\em Proceedings of the IEEE/CVF Conference on Computer Vision and
  Pattern Recognition}, pages 12124--12134, 2022.

\bibitem{dosovitskiy2020image}
Alexey Dosovitskiy, Lucas Beyer, Alexander Kolesnikov, Dirk Weissenborn,
  Xiaohua Zhai, Thomas Unterthiner, Mostafa Dehghani, Matthias Minderer, Georg
  Heigold, Sylvain Gelly, et~al.
\newblock An image is worth 16x16 words: Transformers for image recognition at
  scale.
\newblock {\em arXiv preprint arXiv:2010.11929}, 2020.

\bibitem{d2021convit}
St{\'e}phane d’Ascoli, Hugo Touvron, Matthew~L Leavitt, Ari~S Morcos, Giulio
  Biroli, and Levent Sagun.
\newblock Convit: Improving vision transformers with soft convolutional
  inductive biases.
\newblock In {\em International Conference on Machine Learning}, pages
  2286--2296. PMLR, 2021.

\bibitem{endres2003new}
Dominik~Maria Endres and Johannes~E Schindelin.
\newblock A new metric for probability distributions.
\newblock {\em IEEE Transactions on Information theory}, 49(7):1858--1860,
  2003.

\bibitem{fan2021multiscale}
Haoqi Fan, Bo Xiong, Karttikeya Mangalam, Yanghao Li, Zhicheng Yan, Jitendra
  Malik, and Christoph Feichtenhofer.
\newblock Multiscale vision transformers.
\newblock In {\em Proceedings of the IEEE/CVF International Conference on
  Computer Vision}, pages 6824--6835, 2021.

\bibitem{fei2004learning}
Li Fei-Fei, Rob Fergus, and Pietro Perona.
\newblock Learning generative visual models from few training examples: An
  incremental bayesian approach tested on 101 object categories.
\newblock In {\em conference on computer vision and pattern recognition
  workshop}, pages 178--178. IEEE, 2004.

\bibitem{geiger2013vision}
Andreas Geiger, Philip Lenz, Christoph Stiller, and Raquel Urtasun.
\newblock Vision meets robotics: The kitti dataset.
\newblock {\em The International Journal of Robotics Research},
  32(11):1231--1237, 2013.

\bibitem{han2021transformer}
Kai Han, An Xiao, Enhua Wu, Jianyuan Guo, Chunjing Xu, and Yunhe Wang.
\newblock Transformer in transformer.
\newblock In {\em NeurIPS}, 2021.

\bibitem{he2018soft}
Y He, G Kang, X Dong, Y Fu, and Y Yang.
\newblock Soft filter pruning for accelerating deep convolutional neural
  networks.
\newblock In {\em IJCAI International Joint Conference on Artificial
  Intelligence}, 2018.

\bibitem{helber2019eurosat}
Patrick Helber, Benjamin Bischke, Andreas Dengel, and Damian Borth.
\newblock Eurosat: A novel dataset and deep learning benchmark for land use and
  land cover classification.
\newblock {\em IEEE Journal of Selected Topics in Applied Earth Observations
  and Remote Sensing}, 2019.

\bibitem{hendrycks2021many}
Dan Hendrycks, Steven Basart, Norman Mu, Saurav Kadavath, Frank Wang, Evan
  Dorundo, Rahul Desai, Tyler Zhu, Samyak Parajuli, Mike Guo, et~al.
\newblock The many faces of robustness: A critical analysis of
  out-of-distribution generalization.
\newblock In {\em Proceedings of the IEEE/CVF International Conference on
  Computer Vision (ICCV)}, pages 8340--8349, 2021.

\bibitem{hendrycks2021natural}
Dan Hendrycks, Kevin Zhao, Steven Basart, Jacob Steinhardt, and Dawn Song.
\newblock Natural adversarial examples.
\newblock In {\em Proceedings of the IEEE/CVF Conference on Computer Vision and
  Pattern Recognition (CVPR)}, pages 15262--15271, 2021.

\bibitem{adapter}
Neil Houlsby, Andrei Giurgiu, Stanislaw Jastrzebski, Bruna Morrone, Quentin
  De~Laroussilhe, Andrea Gesmundo, Mona Attariyan, and Sylvain Gelly.
\newblock Parameter-efficient transfer learning for nlp.
\newblock In {\em International Conference on Machine Learning (ICML)}, pages
  2790--2799. PMLR, 2019.

\bibitem{lora}
Edward~J Hu, Yelong Shen, Phillip Wallis, Zeyuan Allen-Zhu, Yuanzhi Li, Shean
  Wang, Lu Wang, and Weizhu Chen.
\newblock Lora: Low-rank adaptation of large language models.
\newblock {\em arXiv preprint arXiv:2106.09685}, 2021.

\bibitem{vpt}
Menglin Jia, Luming Tang, Bor-Chun Chen, Claire Cardie, Serge Belongie, Bharath
  Hariharan, and Ser-Nam Lim.
\newblock Visual prompt tuning.
\newblock In {\em ECCV}, 2022.

\bibitem{jie2022convolutional}
Shibo Jie and Zhi-Hong Deng.
\newblock Convolutional bypasses are better vision transformer adapters.
\newblock {\em arXiv preprint arXiv:2207.07039}, 2022.

\bibitem{johnson2017clevr}
Justin Johnson, Bharath Hariharan, Laurens Van Der~Maaten, Li Fei-Fei, C
  Lawrence~Zitnick, and Ross Girshick.
\newblock Clevr: A diagnostic dataset for compositional language and elementary
  visual reasoning.
\newblock In {\em Proceedings of the IEEE/CVF Conference on Computer Vision and
  Pattern Recognition (CVPR)}, pages 2901--2910, 2017.

\bibitem{kaggle2015retinopathy}
Kaggle and EyePacs.
\newblock Kaggle diabetic retinopathy detection.
\newblock 2015.

\bibitem{krizhevsky2009learning}
Alex Krizhevsky, Geoffrey Hinton, et~al.
\newblock Learning multiple layers of features from tiny images.
\newblock 2009.

\bibitem{kullback1951information}
Solomon Kullback and Richard~A Leibler.
\newblock On information and sufficiency.
\newblock {\em The annals of mathematical statistics}, 22(1):79--86, 1951.

\bibitem{lecun2004learning}
Yann LeCun, Fu~Jie Huang, and Leon Bottou.
\newblock Learning methods for generic object recognition with invariance to
  pose and lighting.
\newblock In {\em Proceedings of the IEEE/CVF Conference on Computer Vision and
  Pattern Recognition (CVPR)}, volume~2, pages II--104. IEEE, 2004.

\bibitem{lepikhin2020gshard}
Dmitry Lepikhin, HyoukJoong Lee, Yuanzhong Xu, Dehao Chen, Orhan Firat, Yanping
  Huang, Maxim Krikun, Noam Shazeer, and Zhifeng Chen.
\newblock Gshard: Scaling giant models with conditional computation and
  automatic sharding.
\newblock {\em arXiv preprint arXiv:2006.16668}, 2020.

\bibitem{li2016pruning}
Hao Li, Asim Kadav, Igor Durdanovic, Hanan Samet, and Hans~Peter Graf.
\newblock Pruning filters for efficient convnets.
\newblock {\em arXiv preprint arXiv:1608.08710}, 2016.

\bibitem{li2016revisiting}
Yanghao Li, Naiyan Wang, Jianping Shi, Jiaying Liu, and Xiaodi Hou.
\newblock Revisiting batch normalization for practical domain adaptation.
\newblock {\em arXiv preprint arXiv:1603.04779}, 2016.

\bibitem{SSF}
Dongze Lian, Daquan Zhou, Jiashi Feng, and Xinchao Wang.
\newblock Scaling \& shifting your features: A new baseline for efficient model
  tuning.
\newblock In {\em Advances in Neural Information Processing Systems (NeurIPS)},
  2022.

\bibitem{liu2022prompt}
Lingbo Liu, Bruce~XB Yu, Jianlong Chang, Qi Tian, and Chang-Wen Chen.
\newblock Prompt-matched semantic segmentation.
\newblock {\em arXiv preprint arXiv:2208.10159}, 2022.

\bibitem{liu2021swin}
Ze Liu, Yutong Lin, Yue Cao, Han Hu, Yixuan Wei, Zheng Zhang, Stephen Lin, and
  Baining Guo.
\newblock Swin transformer: Hierarchical vision transformer using shifted
  windows.
\newblock In {\em Proceedings of the IEEE/CVF International Conference on
  Computer Vision}, pages 10012--10022, 2021.

\bibitem{luo2017thinet}
Jian-Hao Luo, Jianxin Wu, and Weiyao Lin.
\newblock Thinet: A filter level pruning method for deep neural network
  compression.
\newblock In {\em Proceedings of the IEEE international conference on computer
  vision}, pages 5058--5066, 2017.

\bibitem{luo2022channel}
Xu Luo, Jing Xu, and Zenglin Xu.
\newblock Channel importance matters in few-shot image classification.
\newblock In {\em International conference on machine learning}, pages
  14542--14559. PMLR, 2022.

\bibitem{matthey2017dsprites}
Loic Matthey, Irina Higgins, Demis Hassabis, and Alexander Lerchner.
\newblock dsprites: Disentanglement testing sprites dataset, 2017.

\bibitem{molchanov2019importance}
Pavlo Molchanov, Arun Mallya, Stephen Tyree, Iuri Frosio, and Jan Kautz.
\newblock Importance estimation for neural network pruning.
\newblock In {\em Proceedings of the IEEE/CVF conference on computer vision and
  pattern recognition}, pages 11264--11272, 2019.

\bibitem{netzer2011reading}
Yuval Netzer, Tao Wang, Adam Coates, Alessandro Bissacco, Bo Wu, and Andrew~Y
  Ng.
\newblock Reading digits in natural images with unsupervised feature learning.
\newblock 2011.

\bibitem{nie2022pro}
Xing Nie, Bolin Ni, Jianlong Chang, Gaomeng Meng, Chunlei Huo, Zhaoxiang Zhang,
  Shiming Xiang, Qi Tian, and Chunhong Pan.
\newblock Pro-tuning: Unified prompt tuning for vision tasks.
\newblock {\em arXiv preprint arXiv:2207.14381}, 2022.

\bibitem{nilsback2006visual}
M-E Nilsback and Andrew Zisserman.
\newblock A visual vocabulary for flower classification.
\newblock In {\em Proceedings of the IEEE/CVF Conference on Computer Vision and
  Pattern Recognition (CVPR)}, volume~2, pages 1447--1454. IEEE, 2006.

\bibitem{parkhi2012cats}
Omkar~M Parkhi, Andrea Vedaldi, Andrew Zisserman, and CV Jawahar.
\newblock Cats and dogs.
\newblock In {\em Proceedings of the IEEE/CVF Conference on Computer Vision and
  Pattern Recognition (CVPR)}, pages 3498--3505. IEEE, 2012.

\bibitem{raffel2020exploring}
Colin Raffel, Noam Shazeer, Adam Roberts, Katherine Lee, Sharan Narang, Michael
  Matena, Yanqi Zhou, Wei Li, and Peter~J Liu.
\newblock Exploring the limits of transfer learning with a unified text-to-text
  transformer.
\newblock {\em The Journal of Machine Learning Research}, 21(1):5485--5551,
  2020.

\bibitem{rao2021dynamicvit}
Yongming Rao, Wenliang Zhao, Benlin Liu, Jiwen Lu, Jie Zhou, and Cho-Jui Hsieh.
\newblock Dynamicvit: Efficient vision transformers with dynamic token
  sparsification.
\newblock In {\em NeurIPS}, 2021.

\bibitem{recht2019imagenet}
Benjamin Recht, Rebecca Roelofs, Ludwig Schmidt, and Vaishaal Shankar.
\newblock Do imagenet classifiers generalize to imagenet?
\newblock In {\em International Conference on Machine Learning (ICML)}, pages
  5389--5400. PMLR, 2019.

\bibitem{touvron2021going}
Hugo Touvron, Matthieu Cord, Alexandre Sablayrolles, Gabriel Synnaeve, and
  Herv{\'e} J{\'e}gou.
\newblock Going deeper with image transformers.
\newblock In {\em Proceedings of the IEEE/CVF International Conference on
  Computer Vision}, pages 32--42, 2021.

\bibitem{vaswani2017attention}
Ashish Vaswani, Noam Shazeer, Niki Parmar, Jakob Uszkoreit, Llion Jones,
  Aidan~N Gomez, {\L}ukasz Kaiser, and Illia Polosukhin.
\newblock Attention is all you need.
\newblock In {\em NeurIPS}, 2017.

\bibitem{Veeling2018qh}
Bastiaan~S Veeling, Jasper Linmans, Jim Winkens, Taco Cohen, and Max Welling.
\newblock Rotation equivariant cnns for digital pathology.
\newblock In {\em International Conference on Medical image computing and
  computer-assisted intervention}, pages 210--218. Springer, 2018.

\bibitem{wang2020tent}
Dequan Wang, Evan Shelhamer, Shaoteng Liu, Bruno Olshausen, and Trevor Darrell.
\newblock Tent: Fully test-time adaptation by entropy minimization.
\newblock In {\em ICLR}, 2021.

\bibitem{wang2019learning}
Haohan Wang, Songwei Ge, Zachary Lipton, and Eric~P Xing.
\newblock Learning robust global representations by penalizing local predictive
  power.
\newblock In {\em NeurIPS}, 2019.

\bibitem{wang2021kvt}
Pichao Wang, Xue Wang, Fan Wang, Ming Lin, Shuning Chang, Wen Xie, Hao Li, and
  Rong Jin.
\newblock Kvt: k-nn attention for boosting vision transformers.
\newblock {\em arXiv preprint arXiv:2106.00515}, 2021.

\bibitem{wang2022fine}
Shijie Wang, Jianlong Chang, Zhihui Wang, Haojie Li, Wanli Ouyang, and Qi Tian.
\newblock Fine-grained retrieval prompt tuning.
\newblock {\em arXiv preprint arXiv:2207.14465}, 2022.

\bibitem{pvt}
Wenhai Wang, Enze Xie, Xiang Li, Deng-Ping Fan, Kaitao Song, Ding Liang, Tong
  Lu, Ping Luo, and Ling Shao.
\newblock Pyramid vision transformer: A versatile backbone for dense prediction
  without convolutions.
\newblock In {\em Proceedings of the IEEE/CVF international conference on
  computer vision}, pages 568--578, 2021.

\bibitem{wangvtc}
Zhenyu Wang, Hao Luo, WANG Pichao, Feng Ding, Fan Wang, and Hao Li.
\newblock Vtc-lfc: Vision transformer compression with low-frequency
  components.
\newblock In {\em Advances in Neural Information Processing Systems}.

\bibitem{xiao2010sun}
Jianxiong Xiao, James Hays, Krista~A Ehinger, Aude Oliva, and Antonio Torralba.
\newblock Sun database: Large-scale scene recognition from abbey to zoo.
\newblock In {\em 2010 IEEE computer society conference on computer vision and
  pattern recognition}, pages 3485--3492. IEEE, 2010.

\bibitem{xing2022class}
Yinghui Xing, Qirui Wu, De Cheng, Shizhou Zhang, Guoqiang Liang, and Yanning
  Zhang.
\newblock Class-aware visual prompt tuning for vision-language pre-trained
  model.
\newblock {\em arXiv preprint arXiv:2208.08340}, 2022.

\bibitem{yang2021nvit}
Huanrui Yang, Hongxu Yin, Pavlo Molchanov, Hai Li, and Jan Kautz.
\newblock Nvit: Vision transformer compression and parameter redistribution.
\newblock {\em arXiv preprint arXiv:2110.04869}, 2021.

\bibitem{yuan2021tokens}
Li Yuan, Yunpeng Chen, Tao Wang, Weihao Yu, Yujun Shi, Zi-Hang Jiang,
  Francis~EH Tay, Jiashi Feng, and Shuicheng Yan.
\newblock Tokens-to-token vit: Training vision transformers from scratch on
  imagenet.
\newblock In {\em Proceedings of the IEEE/CVF International Conference on
  Computer Vision}, pages 558--567, 2021.

\bibitem{zhai2019large}
Xiaohua Zhai, Joan Puigcerver, Alexander Kolesnikov, Pierre Ruyssen, Carlos
  Riquelme, Mario Lucic, Josip Djolonga, Andre~Susano Pinto, Maxim Neumann,
  Alexey Dosovitskiy, et~al.
\newblock A large-scale study of representation learning with the visual task
  adaptation benchmark.
\newblock {\em arXiv preprint arXiv:1910.04867}, 2019.

\bibitem{zhang2023multimodal}
Bowen Zhang, Xiaojie Jin, Weibo Gong, Kai Xu, Zhao Zhang, Peng Wang, Xiaohui
  Shen, and Jiashi Feng.
\newblock Multimodal video adapter for parameter efficient video text
  retrieval.
\newblock {\em arXiv preprint arXiv:2301.07868}, 2023.

\bibitem{noah}
Yuanhan Zhang, Kaiyang Zhou, and Ziwei Liu.
\newblock Neural prompt search.
\newblock {\em arXiv preprint arXiv:2206.04673}, 2022.

\bibitem{zheng2022prompt}
Zangwei Zheng, Xiangyu Yue, Kai Wang, and Yang You.
\newblock Prompt vision transformer for domain generalization.
\newblock {\em arXiv preprint arXiv:2208.08914}, 2022.

\bibitem{zhou2021elsa}
Jingkai Zhou, Pichao Wang, Fan Wang, Qiong Liu, Hao Li, and Rong Jin.
\newblock Elsa: Enhanced local self-attention for vision transformer.
\newblock {\em arXiv preprint arXiv:2112.12786}, 2021.

\end{thebibliography}
}

\end{document}